\documentclass[pdflatex,sn-mathphys-ay]{sn-jnl}

\usepackage{graphicx}%
\usepackage{multirow}%
\usepackage{amsmath,amssymb,amsfonts}%
\usepackage{amsthm}%
\usepackage{mathrsfs}%
\usepackage[title]{appendix}%
\usepackage{xcolor}%
\usepackage{textcomp}%
\usepackage{manyfoot}%
\usepackage{booktabs}%
\usepackage{algorithm}%
\usepackage{algorithmicx}%
\usepackage{algpseudocode}%
\usepackage{listings}%
\usepackage{multicol}
\usepackage{comment}
\usepackage[most]{tcolorbox}
\usepackage{subcaption}
\usepackage{placeins}   
\usepackage{makecell}
\usepackage{tabularx}
\usepackage{hyperref}

\setcitestyle{aysep={,}}

\begin{document}

\title{A Dual-Process Perspective on Nudge Susceptibility in LLM-Based GUI Agents}

\author*[1]{\fnm{Haya} \sur{Halimeh}}\email{haya.halimeh@uni-paderborn.de}
\author[1]{\fnm{Sascha} \sur{Kaltenpoth}}\email{sascha.kaltenpoth@uni-paderborn.de}
\author[1]{\fnm{Kevin} \sur{Bösch}}\email{kboesch@mail.uni-paderborn.de}
\author[1]{\fnm{Oliver} \sur{Müller}}\email{oliver.mueller@uni-paderborn.de}

\affil[1]{\orgdiv{Information Systems}, \orgname{Paderborn University}, \orgaddress{\city{Paderborn}, \country{Germany}}}


\abstract{LLM-based GUI agents increasingly act on behalf of users in digital environments that were designed with human users in mind. These graphical user interfaces were designed to support, but also deliberately steer, the behaviour and decisions of users. While behavioural biases in the textual outputs of LLMs are well-documented, far less is known about how such influence operates when models act as agents that perceive interfaces and execute decisions---and, in particular, whether the reasoning capabilities increasingly built into these agents make them more robust to it. Drawing on Dual-Process Theory, we empirically investigate whether LLM-based GUI agents are susceptible to automatic (Type 1) and reflective (Type 2) digital nudges, and how their reasoning configuration moderates this susceptibility. In a randomized online shopping experiment with 3,600 agents and a total of 21,600 simulations across six frontier models from three providers, we found that agents were vulnerable to both nudge types. Crucially, the reasoning configuration moderated these effects in opposing directions, reducing susceptibility to automatic default nudges while heightening it to reflective social influence nudges. Extensive reasoning therefore did not make agents more robust but redirected the route through which choice architecture takes effect. Exploratory analysis further showed this redirection to be systematically structured by model scale. Beyond establishing nudge susceptibility as a behavioural property of agentic AI, the study positions interface design as a governance concern for organizations that delegate decisions to autonomous agents.
}

\keywords{Digital Nudging, Dual-Process Theory, Large Language Model, Reasoning, GUI Agents}

\maketitle

\section{Introduction}\label{sec:introduction}

Agentic AI refers to autonomous systems capable of pursuing user-delegated goals with limited human intervention and are increasingly integrated into organizational information systems \citep{bairdNextGenerationResearch2021,murrayHumansTechnologyForms2021,Murugesan2025-xs}. Recent advances in large language models (LLMs) have given rise to agentic AI that pairs LLMs with the ability to perceive and operate graphical user interfaces (GUIs), referred to as LLM-based GUI agents \citep{huOSAgentsSurvey2025,nguyenGUIAgentsSurvey2025,sagerComprehensiveSurveyAgents2025}, to carry out tasks in digital environments. LLM-based GUI agents can navigate websites \citep{zhou2024webarena}, complete spreadsheet tasks \citep{li2023sheetcopilot}, make purchases \citep{yao2022webshop}, and even conduct financial transactions \citep{yuFinMemPerformanceEnhancedLLM2025}. Examples include ChatGPT Agent \citep{openaiChatGPTAgentSystem2025}, Perplexity's Comet browser \citep{perplexityCometBrowser2025}, and OpenClaw \citep{steinbergerIntroducingOpenClaw2026}. This development marks a shift from LLMs as tools for generating and processing information toward systems capable of acting on users' behalf, enabling new forms of joint human--AI agency \citep{zhanglarge,murrayHumansTechnologyForms2021}.

The growing capabilities of agentic AI are accompanied by accelerating organizational interest and adoption. A recent McKinsey survey found that 23\% of firms plan to scale their use of  such systems  \citep{singlaStateAI20252025}, while industry forecasts suggest that they will significantly reshape online shopping by 2030 \citep{faz2026kiagenten,spaene2026agentic}. As these systems continue to mature, their ability to automate increasingly complex tasks is expected to expand \citep{schmidt2026agentic}, motivating organizations to integrate them into more business processes and increasing users' willingness to delegate decisions to them \citep{candrian2022rise}. Consequently, agentic AI is likely to become commonplace across domains such as e-commerce, healthcare, and finance \citep{banihaniAIDecisionmakingProcess2024}.

Such delegation is consequential because LLM-based GUI agents operate in digital environments built for humans, settings designed not only to facilitate decisions but often also to influence them. A prominent example for influencing decisions in such environments are digital nudges \citep{thalerChoiceArchitectureBehavioral2013,weinmannDigitalNudging2016}. The central premise of nudge theory is that human, and potentially agentic, behaviour can be systematically shaped by subtle and seemingly insignificant changes in the choice architecture \citep{thalerNudgeImprovingDecisions2009}, such as a preselected default or a visually highlighted option \citep{weinmannDigitalNudging2016}. The effectiveness of nudges in steering human behaviour is documented across decades of behavioural research and a wide range of domains \citep{bergramDigitalLandscapeNudging2022,caraban23WaysNudge2019,haki2023digital,hummelHowEffectiveNudging2019}.

One might expect LLM-based agents to be less susceptible to such influences than humans, since they arguably lack the cognitive limitations traditionally invoked to explain human bias. For example, researchers have argued that LLMs exhibit higher economic rationality than humans \citep{chen2023emergence}. A growing body of research, however, points the other way. Various studies suggest that LLMs are similarly subject to bounded rationality \citep{simonBehavioralModelRational1955} and can display systematic and sometimes unstable shifts in their choices when exposed to external influence \citep{cherep2024superficial}. Research on LLM textual outputs in particular demonstrates human-like decision biases, including framing, anchoring, and availability effects \citep{nguyen2024human,suri2024large,hagendorff2023human}, and sensitivity to superficial changes in how options are presented \citep{cherep2024superficial,jones2022capturing}, although the resulting deviations do not always mirror human patterns \citep{macmillan2024ir}.

Yet despite rising interest in agentic AI \citep{schmidt2026agentic}, and specifically in LLM-based GUI agents \citep{zhanglarge,huOSAgentsSurvey2025,nguyenGUIAgentsSurvey2025,sagerComprehensiveSurveyAgents2025}, current research on biases of such systems falls short in two respects. 
First, existing work focuses predominantly on biases in LLMs' textual output \citep{tjuatja2024llms,acerbi2023large,echterhoff2024cognitive} and does not connect the biases it documents to the behavioural science from which nudging derives, yielding a fragmented picture that catalogues effects without explaining why particular effects emerge. Nudge theory \citep{thalerNudgeImprovingDecisions2009}, grounded in Dual-Process Theory \citep{kahneman2003maps,evansDualProcessTheoriesHigher2013}, provides such a framework by classifying nudges by the mode of processing they engage, distinguishing automatic Type 1 from reflective Type 2 nudges \citep{hansenNudgeManipulationChoice2013}. \citet{bosch2025biased} shows that the decoy effect carries over to such agents. However, to the best of our knowledge,
it remains an open question, whether the broader Type 1 and Type 2 distinction  manifests similarly in agentic behaviour, in which an LLM-based GUI agent perceives a graphical interface, plans, navigates, and commits to a choice by acting on it \citep{nguyenGUIAgentsSurvey2025,sagerComprehensiveSurveyAgents2025}.
Second, a common assumption is that greater model capability, particularly through enhanced reasoning configurations, produces more robust and less biased behaviour \citep{wei2022emergent,hagendorff2023human,zhang2025system}. These configurations are designed to support more deliberative reasoning and improve task adaptability \citep{huangReasoningLargeLanguage2023,weiChainofThoughtPromptingElicits2022}, yet evidence that they consistently improve robustness remains mixed \citep{mckenzie2023inverse,sharma2024towards}. Since reasoning effort is a configurable parameter in contemporary LLM systems \citep{openaiChatGPTAgentSystem2025,gemini,anthropic}, determining whether increased reasoning reduces susceptibility to manipulative choice architecture is an important empirical question with immediate practical implications, given that users increasingly rely on agent outputs with limited verification \citep{rheu2025trap,steyvers2025large}.

The present study addresses these gaps by asking whether the foundational premise of nudging extends to agentic decision-making beyond textual output, and how reasoning shapes it through the following research questions:

\begin{itemize}
    \item[] \textbf{RQ1.} \textit{Do LLM-based GUI agents exhibit systematic shifts in choice behaviour when exposed to Type 1 (automatic) and Type 2 (reflective) nudges embedded in a graphical choice environment?}
    \item[] \textbf{RQ2.} \textit{How does the reasoning configuration of LLM-based GUI agents moderate their susceptibility to Type 1 (automatic) and Type 2 (reflective) nudges?}
\end{itemize}

To answer these questions, we treat an agent's reasoning configuration as a functional proxy for the two modes of Dual-Process Theory, with a no-reasoning configuration corresponding to automatic System 1-like processing and a high-reasoning configuration to reflective System 2-like processing. This operationalisation builds on a growing body of work that casts explicit, step-by-step reasoning in LLMs as a computational analogue of deliberate System 2 processing \citep{zhang2025system,booch2021thinking,weston2023system}. Rather than assuming this correspondence, we validate it through a manipulation check showing that increasing the reasoning configuration improves performance on a canonical System 2 task (Appendix~\ref{app:reasoningproof}). We use the analogy only at the level of observable behaviour and make no claim that agents possess these cognitive systems \citep{shanahanRolePlayLarge2023}.

Using this operationalisation, we conduct a set of computational simulations that adapt an established human decision-making experiment in an online shopping context \citep{ingendahlWhoCanBe2021} to LLM-based GUI agents, providing empirical insight into how they respond to equivalent nudge interventions \citep{bairdNextGenerationResearch2021}. Concretely, we generate 3,600 agent instances across six leading models that pair each provider's flagship with its smaller variant, namely GPT-5.4 and GPT-5.4-mini \citep{openaiIntroducingGPT52025}, Gemini 3.5 Flash and Gemini 3.1 Flash Lite \citep{gemini}, and Claude Sonnet 4.6 and Claude Haiku 4.5 \citep{anthropic}. Agents are assigned to one of two reasoning configurations and one of three experimental conditions: a control, a Type 1 nudge (default), or a Type 2 nudge (social influence). Each of the 3,600 agents completes a single shopping episode in a simulated grocery store modelled closely on the interface of the original experiment \citep{ingendahlWhoCanBe2021}, selecting one product from each of six categories. Across all agents, this results in 21,600 simulated decisions.

The results reveal that LLM-based GUI agents are not only susceptible to nudge interventions in digital environments, but also show that reasoning configuration and nudge type interact substantially. Consistent with Dual-Process Theory, increasing an agent's reasoning does not eliminate the influence of nudges but shifts the pathway through which they exert their effect.  On average, agents with higher reasoning move from being primarily affected by automatic Type 1 nudges to being more responsive to reflective Type 2 nudges. Per-model and per-provider analyses further reveal substantial but structured heterogeneity. The attenuation of default nudges under high reasoning is concentrated in the flagship models, whereas the amplification of social influence nudges is concentrated in the smaller model variants, suggesting that architectural, training, and scale differences shape the degree to which reasoning modulates nudge susceptibility.

This study makes three contributions. 
First, it advances the debate on bounded rationality in AI decision-making \citep{simonBehavioralModelRational1955,nguyen2024human,chen2023emergence} by moving beyond isolated demonstrations of individual biases \citep{bosch2025biased,cherep2024superficial} to a theory-linked account that explains how reasoning affects agents' susceptibility to nudges. Second, it contributes to theorizing on agentic IT artifacts \citep{bairdNextGenerationResearch2021} by establishing nudge susceptibility as a behavioural property of agentic AI rather than a phenomenon unique to humans, general across model families and, in exploratory analysis, shaped by model scale.
Third, it bridges behavioural economics and information systems research by demonstrating that Dual-Process Theory \citep{kahneman2003maps}, long applied to human decision-makers, can inform the study of autonomous agents that do not reason like humans yet behave in human-like ways within digital choice environments \citep{brady2025dual}. 
In particular, we show that the theory's characteristic behavioural signature---reflective processing moderating automatic influence---arises in decision-makers that possess none of the cognitive architecture or resource constraints the theory assumes, and that engaging deliberation changes susceptibility across nudge types rather than uniformly reducing it. This decouples the dual-process behavioural pattern from its human cognitive substrate and refines the theory's prediction that System 2 engagement debiases judgement.

From a practical perspective, the study highlights the need for caution among organizations and individuals deploying autonomous agents. Since digital choice architectures can materially affect agent decisions, interface design becomes a matter of governance rather than usability alone, and configuring agents for more deliberation changes which architectures influence them rather than whether they do. From an academic perspective, the findings warrant closer investigation of the robustness of AI decision-making in human-designed environments and underscore the need to study agentic behaviour with the same behavioural rigour applied to human decision-making. Otherwise, we risk delegating authority to systems whose choices remain easily swayed by spurious cues.

The remainder of this paper is structured as follows. Section~\ref{sec:background} provides the theoretical background on reasoning in LLMs, LLM-based agents, and nudge theory. Section~\ref{sec:hypotheses} derives the hypotheses from this foundation. Section~\ref{sec:design} outlines the research design, and Section~\ref{sec:results} presents the empirical results. Section~\ref{sec:discussion} discusses the findings and their implications for research and practice. Section~\ref{sec:conclusion} concludes with limitations and directions for future research.

\section{Theoretical Background}\label{sec:background}

\subsection{Reasoning in Large Language Models}\label{subsec:reasoning}

Large Language Models are generative AI models, pre-trained on vast amounts of data and comprising billions of parameters \citep{feuerriegelGenerativeAI2024}. During pre-training, they learn a statistical distribution to auto-regressively predict the next token (word part) based on the previous token sequence \citep{shanahanRolePlayLarge2023}. On this basis, such models are capable of generalizing across diverse tasks without requiring additional task-specific training, relying solely on textual prompts comprising natural language instructions or illustrative few-shot examples \citep{feuerriegelGenerativeAI2024,liuPretrainPromptPredict2023}.

Beyond task instructions, LLMs can also be prompted to perform step-by-step reasoning about the task at hand, known as Chain-of-Thought (CoT) prompting \citep{weiChainofThoughtPromptingElicits2022}. By generating intermediate reasoning steps rather than a direct answer, LLMs emulate human reasoning in textual form, which often facilitates better task adaptation \citep{huangReasoningLargeLanguage2023,weiChainofThoughtPromptingElicits2022}. This approach shifts LLMs away from fast and error-prone responses toward more deliberate and reflective processing \citep{zhang2025system}.

This distinction arguably parallels Dual-Process Theory in behavioural economics \citep{kahneman2003maps,evansDualProcessTheoriesHigher2013}, which states that human judgement relies on two complementary cognitive systems. System 1 operates rapidly, automatically, and intuitively while requiring minimal mental effort, whereas System 2 functions more slowly and analytically by engaging effortful processing when individuals encounter complex or non-routine tasks. We emphasize that the parallel is functional rather than mechanistic, and that the analogy concerns observable input--output behaviour under different processing regimes, not the underlying computation \citep{shanahanRolePlayLarge2023}.

Because LLMs generate outputs by predicting the conditional probability of the next token, they may exhibit System 1-like behaviour, including cognitive biases that manifest as heuristics or shortcuts acquired from training data \citep{jones2022capturing,suri2024large}. Existing work indeed shows that earlier models systematically reproduce the intuitive errors that characterize human System 1 responding \citep{hagendorff2023human,binz2023using}. A prominent example is the bat-and-ball problem \citep[p.~44]{watson2011d}:
\begin{quote}
    ``A bat and ball cost \$1.10. The bat costs one dollar more than the ball. How much does the ball cost?''
\end{quote}
When relying on System 1, humans and earlier LLMs (specifically GPT-3) frequently produce the intuitive but incorrect answer of \$0.10 \citep{binz2023using,hagendorff2023human}. Arriving at the correct solution of \$0.05 requires additional mental effort and engaging System 2 thinking \citep{watson2011d}.

Recent advances in LLM training have introduced mechanisms that encourage more explicit step-by-step reasoning, aiming to reduce susceptibility to System 1-style biases and improve overall performance \citep{hagendorff2023human,zhang2025system}. Models that integrate this process internally, often referred to as reasoning models \citep{huangReasoningLargeLanguage2023,zhang2025system}, generally achieve higher accuracy on complex tasks and appear less vulnerable to certain failures associated with immediate, heuristic-based responding \citep{huangReasoningLargeLanguage2023,zhang2025system}.

Crucially, however, unlike human thinking, in which individuals shift dynamically between System 1 and System 2 to meet the demands of the situation \citep{kahneman2003maps}, LLMs do not switch between these modes autonomously in the same way. Instead, the degree of reasoning is typically determined by their training and configuration \citep{zhang2025system}. Current models permit reasoning effort to be adjusted programmatically, ranging from none to high \citep{openaiChatGPTAgentSystem2025,gemini,anthropic}. Non-reasoning configurations resemble System 1 in that they favour fast and less deliberative responses, whereas high-reasoning configurations resemble System 2 in that they involve slower and more effortful processing.

Nevertheless, the evidence that stronger reasoning yields more robust behaviour is mixed. While higher reasoning can reduce biases overall, and System 1 biases in particular \citep{zhang2025system,zhanglarge}, high-reasoning models remain sensitive to systematic errors \citep{huangReasoningLargeLanguage2023,guo2025illusion}, and their deviations do not always mirror human ones \citep{macmillan2024ir}. Increased scale can even worsen performance on some tasks, a pattern termed inverse scaling \citep{mckenzie2023inverse}, and extended chains of reasoning can degrade decisions where deliberation is counterproductive \citep{liu2025mind}. Critically, greater capability and stronger alignment to human preferences can introduce new vulnerabilities rather than remove existing ones. A salient example is sycophancy, the tendency of models trained with human feedback to conform to a user's stated views, which becomes more pronounced in larger and more heavily aligned models \citep{perez2023discovering,sharma2024towards}. A closely related phenomenon is conformity, whereby LLMs align with a presented majority even when it is incorrect \citep{weng2025we}. Reduced cognitive bias therefore does not necessarily imply more robust decision-making \citep{DentellaDisplay}. This matters especially for LLM-based GUI agents, where reasoning may shift the route of external influence rather than close it off.

\subsection{Large Language Model-based Agents}\label{subsec:agents}

Beyond enhanced reasoning capabilities, LLMs can be augmented to use external tools such as search engines, calculators, calendars, or application programming interfaces (APIs) \citep{quToolLearningLarge2025,schickToolformerLanguageModels2023}. When an LLM combines tool use with reasoning, it adopts the reasoning-and-acting (ReAct) framework, which enables it to operate as an autonomous agent that interleaves deliberation with action \citep{goldiMakingSenseLarge2025}.
Building on this capability, LLM-based agents can now interact with graphical user interfaces (GUIs) such as operating systems, software applications, and browsers \citep{huOSAgentsSurvey2025,nguyenGUIAgentsSurvey2025,sagerComprehensiveSurveyAgents2025}. These systems, often referred to as computer control agents \citep{sagerComprehensiveSurveyAgents2025} or GUI agents \citep{nguyenGUIAgentsSurvey2025}, move beyond conversational exchange to autonomous action. A GUI agent perceives the interface through rendered pixels and structured representations such as the document object model, plans over multiple steps, navigates between elements, and commits to a choice by taking an action on the user's behalf \citep{nguyenGUIAgentsSurvey2025,sagerComprehensiveSurveyAgents2025}. This perception-plan-act loop distinguishes agentic behaviour from the single-turn text elicitation in which LLM decision-making has typically been studied.

Such capabilities underpin products such as Perplexity's Comet browser \citep{perplexityCometBrowser2025}, ChatGPT Agent \citep{openaiIntroducingChatGPTAgent2025}, and open-source projects like the Browser Use library \citep{browseruseteamCelebratingOneYear2025}, which in turn open the door to large-scale task automation \citep{marreedEnterpriseReadyComputerUsing2025}. As these agents become capable of automating information retrieval, product identification, and purchasing, organizations are increasingly embedding them into business processes \citep{singlaStateAI20252025}, while users are placing growing expectations on them to autonomously complete tasks on their behalf \citep{perplexityteamInternetBetterComet2025}.
In keeping with this focus, research on  LLM-based GUI agents has concentrated on capability and task success, evaluated through benchmarks of perception, grounding, planning, and goal completion \citep{huOSAgentsSurvey2025,nguyenGUIAgentsSurvey2025,sagerComprehensiveSurveyAgents2025}, with comparatively less attention to how these agents decide when the environment is designed to influence choice.
LLM-based GUI agents are exposed to such influence on two fronts. They inherit the behavioural biases of the LLMs that power them, as discussed in Section~\ref{subsec:reasoning}, and they act within the human-oriented choice architectures embedded in the environments they navigate.

\subsection{Nudging}\label{subsec:nudging}

Grounded in the realization that human judgement suffers from systematic cognitive biases \citep{kahneman2003maps,tversky1974judgment,tversky1981framing}, nudge theory marked a shift in understanding how the design of choice architecture influences human decision-making \citep{hummelHowEffectiveNudging2019}. At its core, a nudge refers to a subtle modification in how options are presented that predictably influences behaviour without forbidding any options, restricting freedom of choice, or significantly altering economic incentives \citep{thalerNudgeImprovingDecisions2009}. It challenges the classical notion of the consistently rational homo economicus, holding that human behaviour often departs from deliberative reasoning and is instead influenced by systematic biases \citep{stanovichRationalityReflectiveMind2011} and bounded rationality \citep{simonBehavioralModelRational1955}.
Over time, the principles of nudging extended from analogue contexts into the digital sphere, giving rise to the notion of digital nudging \citep{weinmannDigitalNudging2016,mirsch2017digital}. Digital nudging steers users toward particular selections within digital choice environments by modifying interface design elements \citep{schneiderDigitalNudgingGuiding2018}, for example through default settings or the visual highlighting of specific options \citep{weinmannDigitalNudging2016}.

Building on Dual-Process Theory \citep{kahneman2003maps,evansDualProcessTheoriesHigher2013}, \citet{hansenNudgeManipulationChoice2013} proposed categorizing nudges according to the mode of thinking they engage. Type 1 nudges act on the automatic system without invoking reflective thinking, whereas Type 2 nudges target the premises and attention of reflective thinking. The two nudges examined in this study, defaults and social influence, are canonical instances of these respective types.

A default nudge preselects one option so that it is realized unless the decision-maker actively chooses otherwise. Its influence operates through the automatic system, which makes it a Type 1 nudge. Defaults exploit the status quo bias, the tendency to prefer an existing or preselected state and to avoid the cognitive effort of evaluating alternatives \citep{samuelson1988status,ritovStatusquoOmissionBiases1992}. Because retaining the preselected option requires no deliberation and follows the path of least resistance, defaults steer behaviour without engaging reflective thought \citep{thalerNudgeImprovingDecisions2009}. The effect is among the most robust in the nudging literature, as illustrated by the finding that specifying organ donation as the default markedly raises participation relative to an opt-in arrangement \citep{johnson2003defaults}.

A social influence nudge presents information about the choices of others, so that the decision-maker adjusts behaviour in light of that reference point. Because processing this information requires attention to and reflection on social cues, it constitutes a Type 2 nudge. Its influence rests on two conformity motives, namely normative influence, the desire to gain social approval, and informational influence, the assumption that the behaviour of others signals the appropriate course of action, particularly under uncertainty \citep{deutschStudyNormativeInformational1955,cialdiniSocialInfluenceSocial1998}. Descriptive social norms operationalise this mechanism by indicating what most people do in a given situation \citep{cialdini1990focus}. Such norms reliably shift behaviour across domains, from towel reuse among hotel guests \citep{goldstein2008room} to household energy conservation \citep{allcott2011social} and pro-environmental product choices in online shopping \citep{demarqueNudgingSustainableConsumption2015}.

Unlike adversarial exploits such as prompt injection, which hide hostile instructions in the content an agent processes to subvert its behaviour \citep{greshake2023not,zhan2024injecagent,debenedetti2024agentdojo}, nudges are a routine feature of online interface design \citep{schneiderDigitalNudgingGuiding2018,caraban23WaysNudge2019}, so agents encounter them by default in the ordinary course of a task rather than as an external intrusion. Because the agent commits to choices on the user's behalf, the consequences of any such influence accrue to the delegating user.

Existing evidence of nudge-like effects in LLMs derives from prompt-level textual elicitation or addresses only a single nudge type. A nudge embedded in a graphical interface, however, must be perceived within the rendered choice environment, integrated across planning steps, and acted upon, so a bias observed in single-turn textual elicitation need not persist through this perception-plan-act loop. Moreover, reasoning effort can be fixed closer to System 1- or System 2-like processing, which makes it a natural moderator of susceptibility, yet how this configuration shapes responsiveness to each nudge type has not yet been examined. The present study addresses these gaps, which the next section develops into testable hypotheses.

\section{Hypothesis Development}\label{sec:hypotheses}

A growing body of research demonstrates that LLMs reproduce human-like cognitive biases in their textual outputs. Models exhibit established behavioural effects such as framing and availability biases \citep{jones2022capturing,hagendorff2023human,suri2024large}, and instruction tuning and alignment can amplify such tendencies rather than eliminate them \citep{itzhak2024instructed,echterhoff2024cognitive}. Beyond individual biases, LLMs display inconsistent decision patterns \citep{guo2025illusion}, exhibit systematic biases when evaluating alternatives \citep{koo2024benchmarking}, and depart from rational-choice models in ways that do not always match human behaviour \citep{macmillan2024ir}, though in applied settings their errors can resemble those of human decision-makers \citep{chen2025manager}.

Several of these vulnerabilities arise through mechanisms closely related to digital nudging. LLMs are sensitive to superficial variations in how options are presented \citep{cherep2024superficial} and reproduce established choice-architecture effects such as anchoring \citep{nguyen2024human}. Most relevant to the present study, \citet{bosch2025biased} show that the decoy effect systematically influences the product choices of LLM-based GUI agents, providing initial evidence that behavioural effects observed in text-based interaction can persist when LLMs act within graphical interfaces.

Taken together, these findings indicate that although LLM-based agents do not ``reason'' in a human sense but generate behaviour through learned statistical patterns \citep{brady2025dual}, their decisions can nonetheless display vulnerabilities comparable to those observed in human cognition. Extrapolating these findings to agentic settings, we expect LLM-based GUI agents operating in online shopping environments to be susceptible to both Type 1 and Type 2 nudges embedded in the interface, each of which reliably affects human choices \citep{johnson2003defaults,goldstein2008room}. 

\textbf{H1:} \textit{LLM-based GUI agents exhibit systematic shifts in choice behaviour when exposed to Type 1 (automatic) and Type 2 (reflective) nudge interventions.}

In human decision-making, the moderating role of cognitive engagement is contested. Some studies find that individual differences in cognitive ability and disposition shape susceptibility to nudges \citep{ingendahlWhoCanBe2021,deridderNudgeabilityMappingConditions2022}, whereas others report that people are similarly nudgeable regardless of the cognitive system engaged \citep{vangestelNudgesMakeUse2021}. To examine whether this relationship extends to LLM-based agents, we treat the reasoning configuration as a functional proxy for the two dual-process modes established in Section~\ref{subsec:reasoning}, with a no-reasoning configuration corresponding to System 1-like and a high-reasoning configuration to System 2-like processing \citep{zhang2025system}. In dual-process terms, deliberation counteracts the automatic responses that Type 1 nudges exploit \citep{stanovichRationalityReflectiveMind2011}.

Type 1 nudges operate through automatic processing and the path of least resistance, without engaging reflection. Because System 2-style reasoning counteracts the intuitive, heuristic-driven responses associated with automatic processing \citep{stanovichRationalityReflectiveMind2011}, an agent under a high-reasoning configuration should weigh the alternatives more deliberately rather than accepting a preselected option by default. We therefore expect a high-reasoning configuration to attenuate susceptibility to automatic nudges relative to a no-reasoning configuration.

\textbf{H2a:} \textit{LLM-based GUI agents under a high-reasoning configuration are less susceptible to Type 1 (automatic) nudges than those under a no-reasoning configuration.}

Type 2 nudges, by contrast, operate by supplying information that the decision-maker is expected to weigh, and thus take effect only if that information is actually processed \citep{caraban23WaysNudge2019}. An agent under a high reasoning configuration, which engages the reflective processing these nudges target, should attend to and integrate such cues more thoroughly than one under a no-reasoning configuration. Greater reasoning may therefore not immunize agents against influence but instead deepen their engagement with reflective cues, increasing rather than reducing susceptibility to Type 2 nudges \citep{brady2025dual}.

\textbf{H2b:} \textit{LLM-based GUI agents under a high-reasoning configuration are more susceptible to Type 2 (reflective) nudges than those under a no-reasoning configuration.}

\section{Research Design}\label{sec:design}

To test our hypotheses, we replicate an established consumer choice experiment by \citet{ingendahlWhoCanBe2021}, which investigated two of the most widely studied nudges, default options and social influence, in an online shopping context. We selected this experiment for three reasons. First, both nudge types have consistently been shown to influence consumer behaviour \citep{demarqueNudgingSustainableConsumption2015,szekely2016nudging} and rank among the most frequently applied nudges in the literature \citep{hummelHowEffectiveNudging2019}. Second, as established in Section~\ref{subsec:nudging}, the two nudges map cleanly onto the dual-process typology, with the default nudge operating as a Type 1 (automatic) intervention and the social influence nudge as a Type 2 (reflective) one \citep{caraban23WaysNudge2019}. Third, the online shopping scenario closely mirrors the digital environments that LLM-based GUI agents are increasingly deployed to navigate.

We adopt the original procedure, materials, and task structure but apply them for a different research objective. Whereas \citet{ingendahlWhoCanBe2021} examined how individual personality traits, such as need for cognition and uniqueness, moderate the effectiveness of these nudges in human consumers, we use the design to investigate how LLM-based GUI agents under different reasoning configurations respond to them.

Before the main experiment, we validated the System 1 versus System 2 interpretation of the reasoning configuration in a pre-experiment built on Kahneman's example of counting the occurrences of a letter as a canonical System 2 task \citep{kahneman2011thinking}, which LLMs solve reliably only when explicit reasoning is engaged \citep{zhang2024counting,xu2025llm}. Across the same six models and reasoning configurations as the main experiment (five letter-counting items, 100 repetitions per configuration), accuracy rose from 80.3\% under no reasoning to 99.3\% under high reasoning, with the gains concentrated on novel pseudo-words that cannot be answered from memory, justifying the reasoning configuration as a functional proxy for System 1 versus System 2 processing. Appendix~\ref{app:reasoningproof} reports the full design, the statistical tests, and a reasoning-trace analysis.

\subsection{Materials}\label{subsec:materials}

\begin{figure}[ht]
\centering
\includegraphics[width=\textwidth]{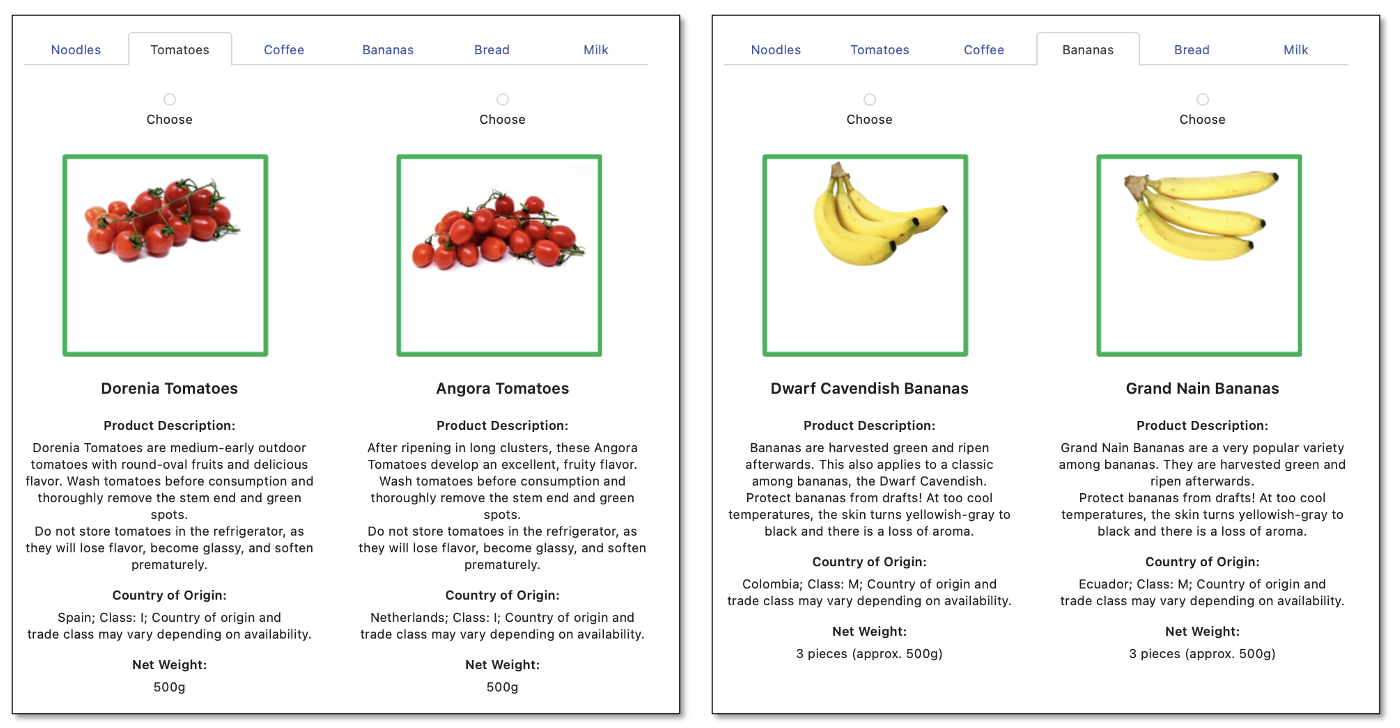}
\caption{Example product display of the simulated grocery store across two categories.}\label{fig:materials}
\end{figure}

We used a simulated online grocery store modelled closely after the interface used in the original human experiment \citep{ingendahlWhoCanBe2021}. The store consisted of six product categories, namely tomatoes, bananas, bread, coffee beans, milk, and pasta. Each category contained two products displayed side by side, each with a product image and a brief description. All original materials, including instructions and descriptions, were translated from German, the language of the original experiment, to English to better align with the linguistic capabilities of the GUI agents. Figure~\ref{fig:materials} shows a representative screenshot of the layout of the store interface.

\subsection{Technical Implementation}\label{subsec:technical}

We implemented the online grocery store using the FastAPI\footnote{\url{https://fastapi.tiangolo.com}} web framework to provide a browser-accessible front end for the experimental task. The LLM-based GUI agents interacted with the store through the Browser Use library \citep{browseruseteamCelebratingOneYear2025}, which enables agents to operate web browsers autonomously by navigating webpages, selecting elements, clicking buttons, entering inputs, reading visual and textual content, and following the web application's own workflow. Through this interface, the agents perceived and interacted with the graphical user interface of the store much as a human participant would. The Browser Use library supports built-in logging, which provides insight into the reasoning behind each agent decision without modifying the system prompt or the instructions given to the agents. To enable reproducibility and future replication, we make the source code and experimental analyses publicly available online\footnote{\url{https://anonymous.4open.science/r/Nudging_LLM_Agents-B1D5}}.

\subsection{Procedure}\label{subsec:procedure}

LLM-based GUI agents were tasked with purchasing six products from the online store for an upcoming meal with friends. At the beginning of each trial, all agents received a standardized prompt that closely mirrored the scenario description used in the original study. The prompt instructed them to visit the online shop, inspect the available products, and follow the on-screen instructions to complete the purchase.

\begin{tcolorbox}[
    enhanced,
    colback=white,
    colframe=black,
    boxrule=0.3pt,
    arc=1mm,
    left=2mm,right=2mm,top=2mm,bottom=2mm,
    title=Prompt,
    attach boxed title to top left={
        xshift=2mm,
        yshift*=-\tcboxedtitleheight/2
    },
    boxed title style={
        colback=white,
        size=small,
        frame hidden,
    },
    fonttitle=\bfseries\color{black},
]
\small
I'm planning a meal for my friends tomorrow. After some quick consideration, I realize I'm still missing a few products.\\[0.3em]
1. Go to the online supermarket \texttt{http://127.0.0.1:8000/}\\
2. Learn about the different products and select your preferred product in each category.\\
3. Follow the instructions on the website carefully to complete your purchase.
\end{tcolorbox}

\begin{figure}[ht]
\centering
\includegraphics[width=\textwidth]{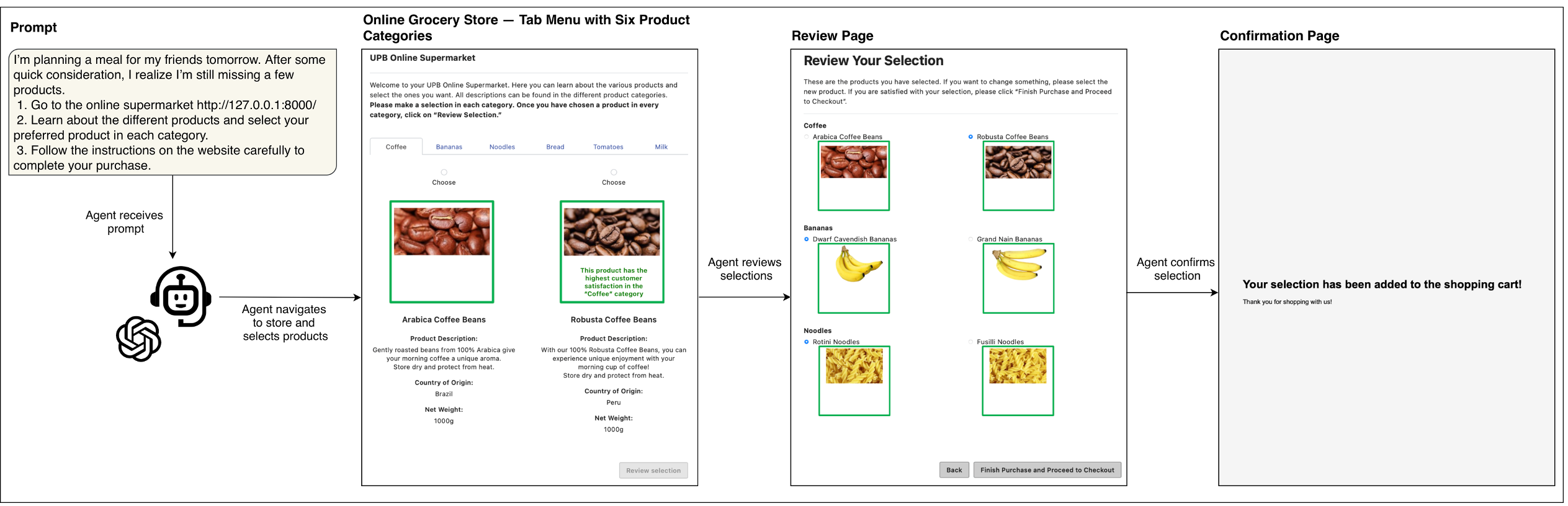}
\caption{The experimental procedure for all simulations.}\label{fig:experiment}
\end{figure}

After receiving the prompt, agents navigated to the online shop, where they browsed the six product categories via a tab menu and selected their preferred product in each category (exactly one of the two options). Once a product from each category was selected, agents advanced to a summary page, where they could review their selections and, if desired, revise them before confirming. The confirmed final selections constituted the dependent variable in our agent-based replication. Upon confirming their selections, agents were redirected to an end page indicating that the items had been added to the shopping cart. This page signalled the completion of the task and allowed the agents to terminate the session and close the online store window. Figure~\ref{fig:experiment} illustrates the full procedure.

\subsection{Conditions}\label{subsec:conditions}

\begin{figure}[ht]
\centering
\includegraphics[width=\textwidth]{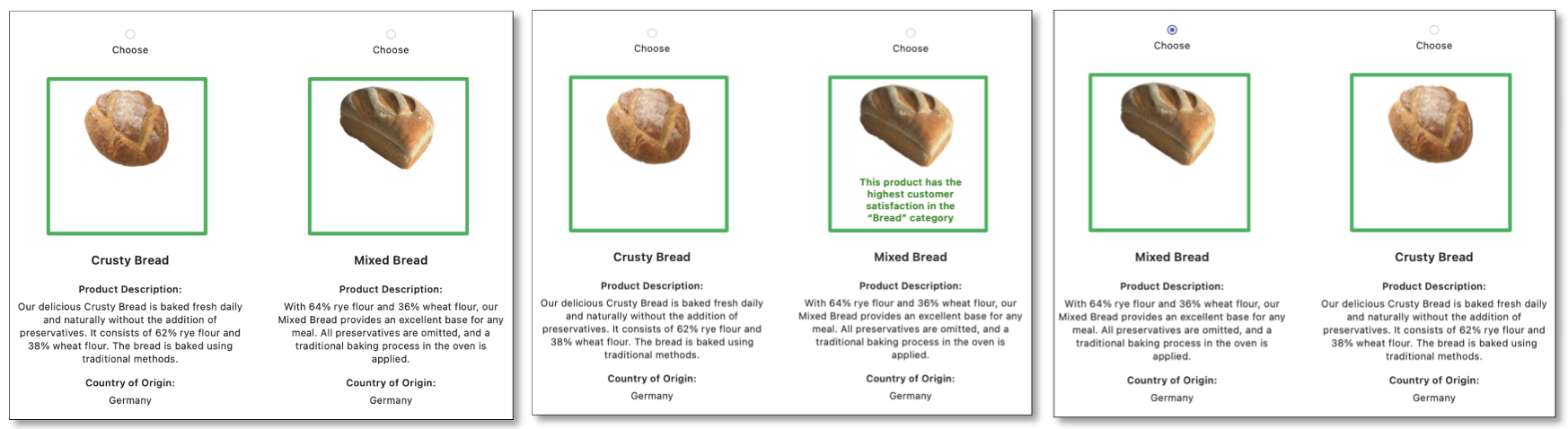}
\caption{The three nudge conditions. The control condition omits both nudge types (left), the social influence nudge condition (center), and the default nudge condition (right).}\label{fig:nudges}
\end{figure}

The experiment followed a three-group between-subjects design. Agents were randomly assigned to one of three conditions. In the control (no-nudge) condition, products were displayed neutrally without any nudge intervention. In the default nudge condition, one of the two products within each category was preselected when the agent opened the page. In the social influence nudge condition, one product per category was presented as having the highest customer satisfaction rate for that category. Figure~\ref{fig:nudges} illustrates the three conditions.

As in the original experiment, both the order of product categories and the position of products within each category were randomized for each agent to mitigate ordering effects. To control for effects of product appearance or description, we employed a standard counterbalanced design. Within each experimental condition, half of the agents were nudged toward one option (product set A) and the other half toward the alternative option (product set B). This design ensures that any inherent advantage of one product set over another cancels out, so that in the absence of a nudging effect the nudged option should be selected at chance level of 50\%.

\subsection{Agent Participants}\label{subsec:sample}

To capture variation across models, we varied the agents' underlying backbone across six leading models from three providers, pairing each provider's flagship model with its smaller, faster variant, namely OpenAI's GPT-5.4 and GPT-5.4-mini \citep{openaiIntroducingGPT52025}, Google's Gemini 3.5 Flash and Gemini 3.1 Flash Lite \citep{gemini}, and Anthropic's Claude Sonnet 4.6 and Claude Haiku 4.5 \citep{anthropic}. This two-factor sampling of backbones (provider $\times$ model size) allows us to assess whether nudge susceptibility and its moderation by reasoning generalize across model families and scales.

In total, we generated 3,600 distinct agents, each conceptually modelled after a human participant, and initialized with one of six LLMs and one of two predefined reasoning configurations. This yielded 600 agents per LLM, evenly split between the reasoning configurations (300 each). We operationalised reasoning by adjusting each model's reasoning effort, yielding two configurations akin to System 1 and System 2 processing, namely no reasoning (reasoning disabled) and high reasoning (high reasoning effort).

Following a between-subjects design, 200 agents per model were assigned to each experimental condition (no nudge control, default nudge, and social influence nudge). Within each condition, agents were evenly divided between the no-reasoning and high-reasoning configurations, with 100 agents in each subgroup. Each agent completed one full shopping episode, selecting one product from each of the six categories. Agents were randomly assigned to receive nudges toward either product set A or product set B. Because each agent interacted once with each product category, this yielded six interactions per trial, 3,600 interactions per model, and 21,600 interactions across all six models.

\subsection{Model Specification}\label{subsec:model}

To test our hypotheses, we estimated Bayesian multilevel logistic regression models (Bernoulli likelihood, logit link) using the brms package (v2.23.0) in R \citep{JSSv080i01}. The unit of analysis was the individual agent interaction, corresponding to one product selection made by an agent during a trial. The dependent variable is a binary indicator of whether the agent selected the product designated as the target, coded 1 if the target option was chosen and 0 otherwise. In the default and social influence conditions, the target corresponded to the nudged option, whereas in the no-nudge condition it was assigned randomly to enable comparability across conditions.

The main independent variables were the experimental condition (control, default, or social influence) and the reasoning configuration of the agents (no reasoning versus high reasoning). All predictors were coded as binary dummy variables. Concretely, the dummy variable for reasoning configuration was coded 0 for no reasoning and 1 for high reasoning, the dummy for the default condition 0 for no default and 1 for default, and the dummy for the social influence condition 0 for no social influence and 1 for social influence. The reference category therefore corresponds to the control condition with no-reasoning agents. To account for the non-independence of repeated choices made by the same agent across the six product categories, we included random intercepts for the agents.

We estimated the models within a Bayesian framework because several design cells, most notably for the two Claude models, exhibited target selection rates at or near 100\%. Such (quasi-)complete separation renders maximum-likelihood estimation of frequentist mixed-effects models (e.g., \textit{glmer} from the lme4 package \citep{batesFittingLinearMixedEffects2015}) unreliable, producing divergent coefficients and standard errors. In the Bayesian specification, a weakly informative Student-$t(3, 0, 2.5)$ prior on all regression coefficients \citep{gelman2008weakly} keeps the posterior proper under separation while leaving all estimable effects essentially unshrunk. All remaining parameters (the intercept and the random-intercept standard deviation) use the brms weakly informative default priors. Each model was estimated with 4 chains of 4,000 iterations (1,000 warmup), yielding 12,000 post-warmup draws, and all fits converged ($\widehat{R} \approx 1.00$, with high effective sample sizes). Because Bayesian models report no $p$-values, we consider an effect \emph{credible} when the 95\% credible interval (CI) of its posterior excludes zero on the log-odds scale, equivalently when the CI of the odds ratio excludes one. For cells at the 100\% ceiling, the point estimate is partly prior-dependent, so conclusions there are read from the data-driven CI bound rather than the point estimate.

Our main analysis comprises three pooled models estimated over all 21,600 interactions of the 3,600 agents. Model~1 regresses target selection on condition, reasoning, and their interaction, with random intercepts per agent ($\text{target} \sim \text{condition} \times \text{reasoning} + (1 \mid \text{agent})$). Model~2 adds the LLM provider (OpenAI, Google, Anthropic, with Anthropic as the reference category) as a fixed-effect control, and Model~3 additionally adds model size (flagship versus small variant, with flagship as the reference). Both are included to verify that the condition and interaction effects are robust to differences between model families and scales. We treat provider purely as a control, since it bundles architecture, training, and alignment differences that cannot be interpreted individually, whereas model size represents scale, a construct we return to in the exploratory analysis in Section~\ref{sec:results}. 

Hypothesis tests are based on posterior odds ratio contrasts derived from the fitted models, namely pairwise contrasts between conditions (H1) and contrasts of high versus no reasoning within each condition (H2a and H2b), summarized by posterior medians with 95\% highest-posterior-density intervals. As robustness checks, we re-estimated the Model~1 specification separately for each provider and for each of the six model backbones. We additionally conducted an exploratory analysis by model size tier (the three flagship and the three small models pooled) to characterize how the reasoning moderation varies with scale. Complete posterior summaries of all models are reported in Appendices~\ref{app:pooled}--\ref{app:permodel}.

\section{Results}\label{sec:results}
\subsection{Hypothesis Tests}\label{subsec:res-hyp}

\begin{figure}[ht]
\centering
    \centering
    \includegraphics[width=\textwidth]{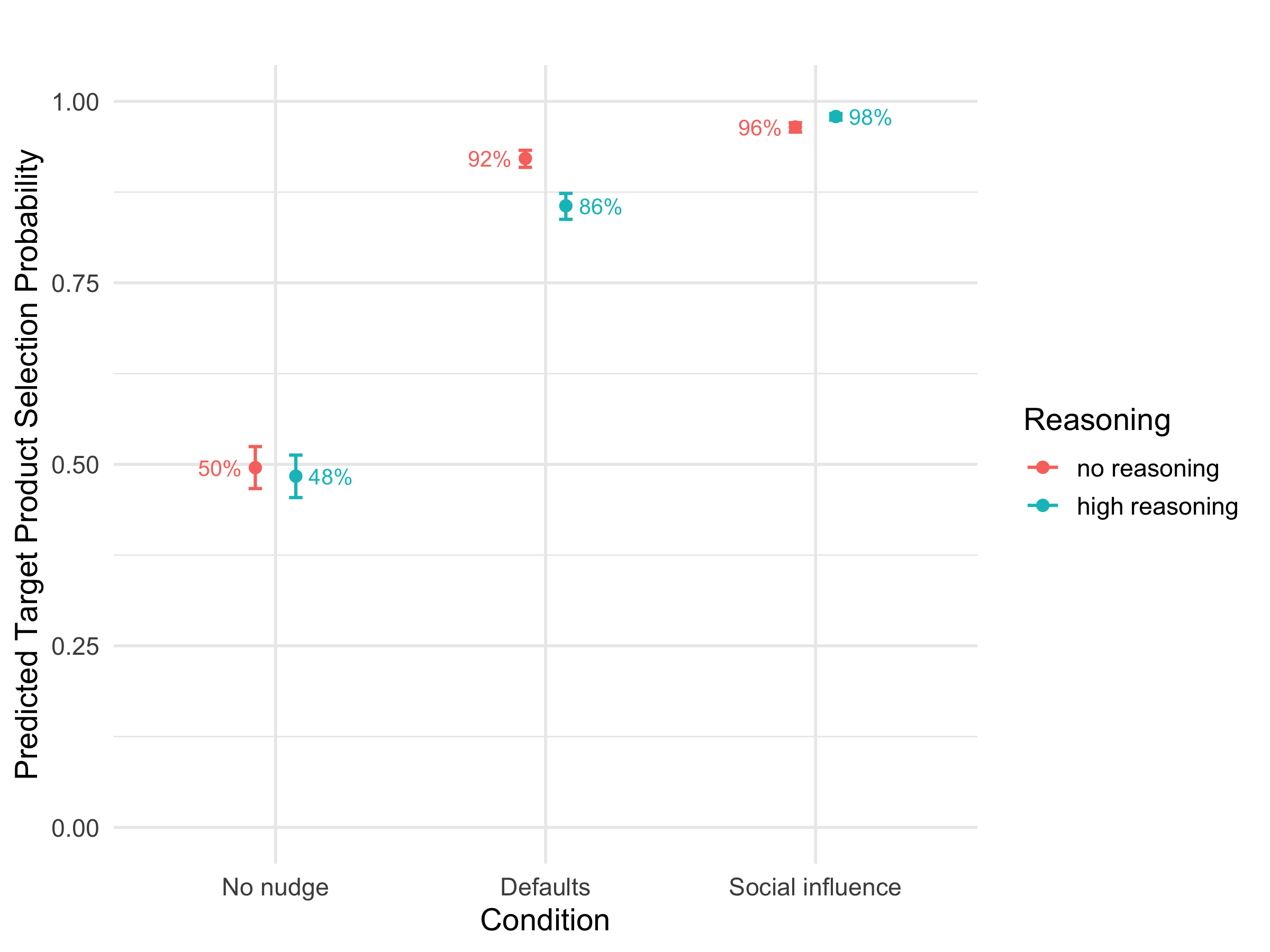}
    \caption{Model 1 (pooled): Estimated target product selection probabilities across conditions and reasoning configurations for the three pooled models. }
    \label{fig:results_m1}
\end{figure}

Figure~\ref{fig:results_m1} displays the estimated probability of selecting the target product for each experimental condition and reasoning configuration, derived from the three pooled Bayesian multilevel logistic regression models. The target is the product toward which the agent is nudged in the treatment conditions. Each panel shows posterior point estimates with 95\% credible intervals. Values above 50\% indicate a systematic shift toward the nudged product.

\begin{figure}
    \centering
    \includegraphics[width=\textwidth]{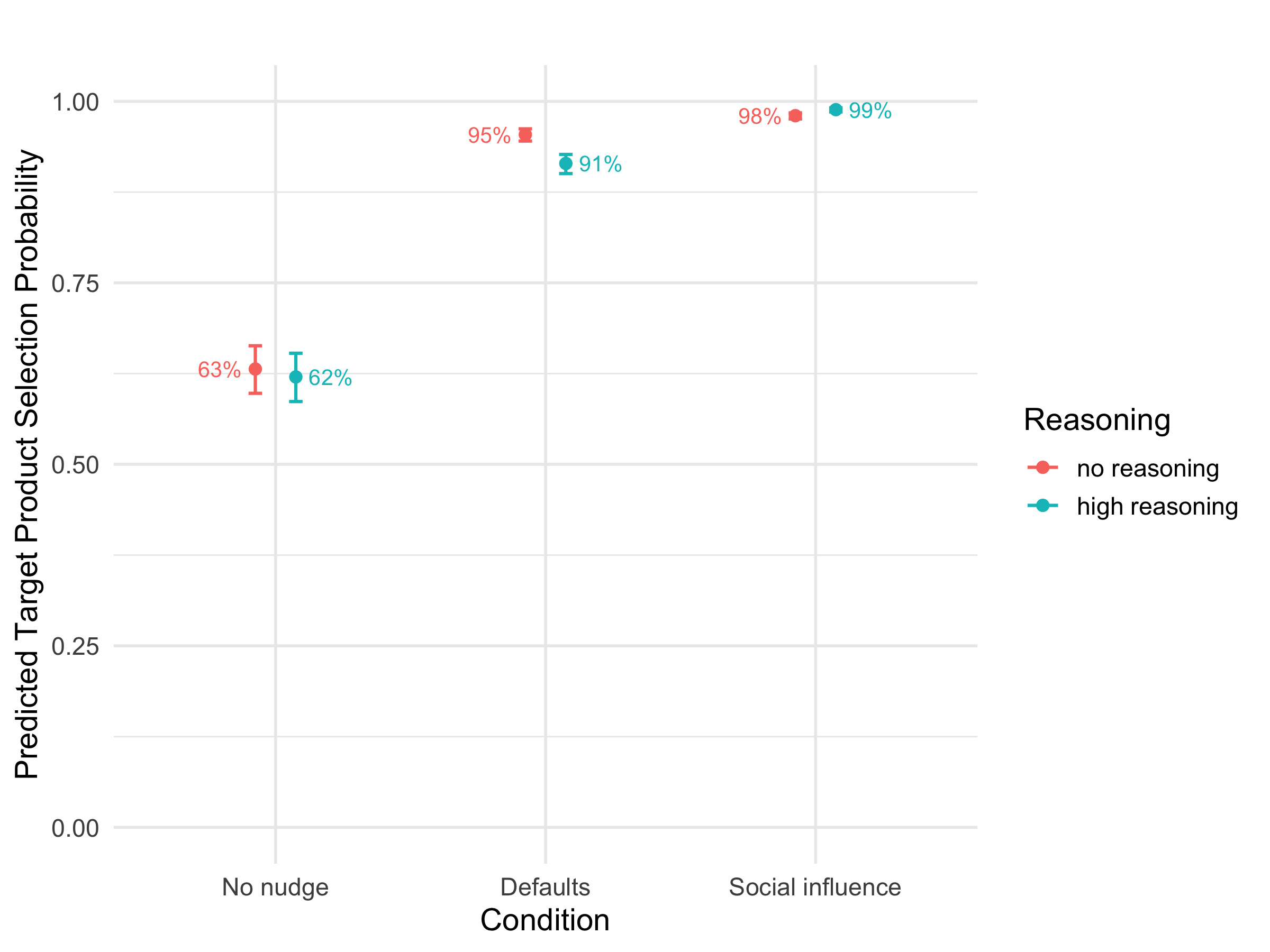}
    \caption{Model 2 (+ provider): Estimated target product selection probabilities across conditions and reasoning configurations for the three pooled models.}
    \label{fig:results_m2}
\end{figure}

\begin{figure}
    \centering
    \includegraphics[width=\textwidth]{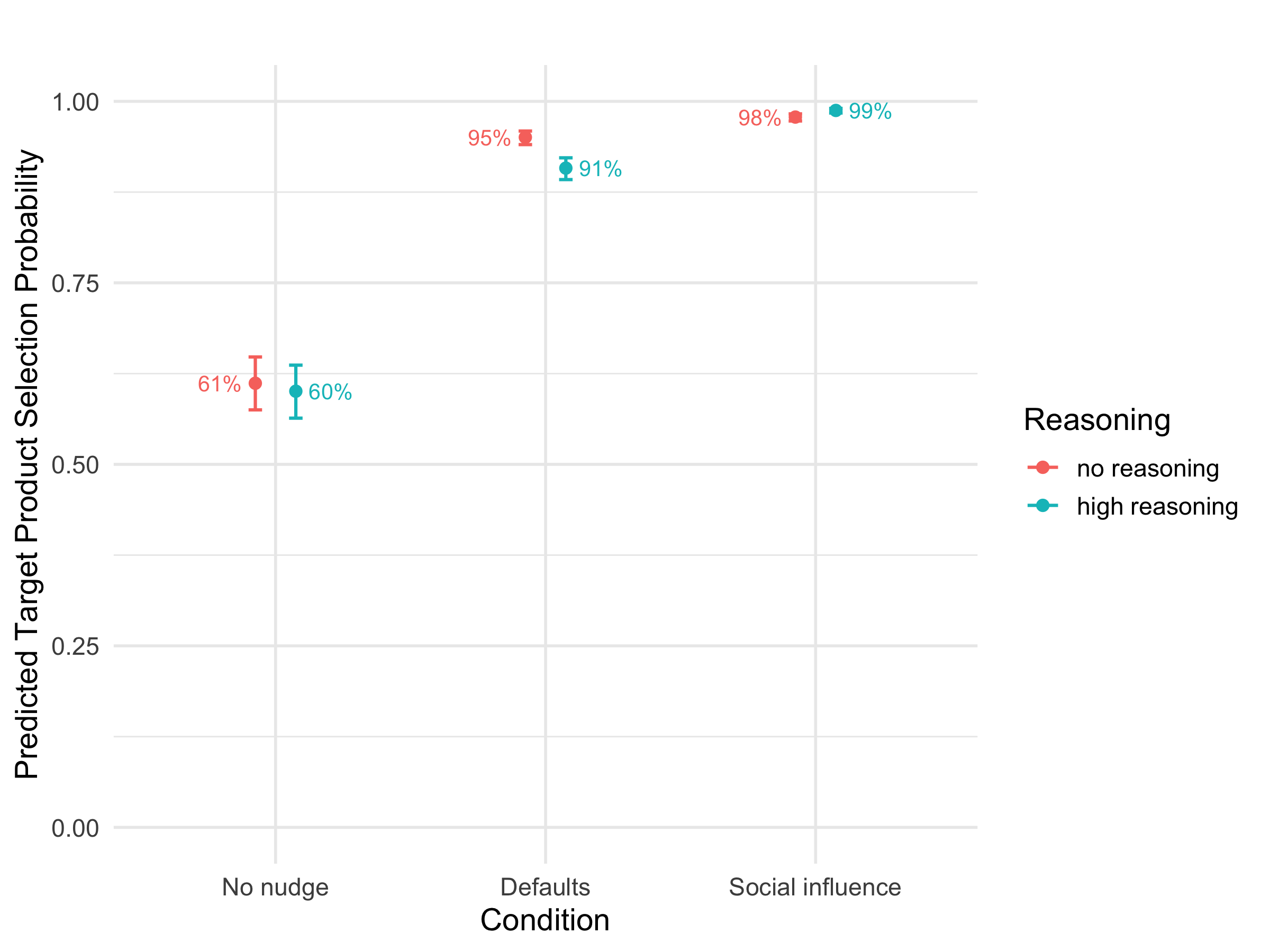}
    \caption{Model 3 (+ provider + size): Estimated target product selection probabilities across conditions and reasoning configurations for the three pooled models.}
    \label{fig:results_m3}
\end{figure}

Looking at the pooled results across all six models (Figure~\ref{fig:results_m1}), several patterns emerge. First, in the control condition without a nudge, the estimated probability of selecting the target product is approximately 50\% ($CI_{.95}$ [47\%, 53\%] for no-reasoning agents and $CI_{.95}$ [46\%, 51\%] for high-reasoning agents), consistent with the counterbalanced design. Because each product serves as the nudge target for half of the agents, inherent preference differences between the two products cancel out by construction, and the expected control-condition selection rate is 50\% at every level of aggregation. Departures from 50\% in the nudge conditions can therefore be attributed to the nudges rather than to the products themselves. 
Second, the estimated mean probability in the default nudge condition rises to approximately 92\% ($CI_{.95}$ [91\%, 93\%]) for no-reasoning agents and 86\% ($CI_{.95}$ [84\%, 87\%]) for high-reasoning agents; here high reasoning \emph{lowers} selection of the preselected option, so agents that reasoned more were less likely to accept the default. 
Third, in the social influence nudge condition the probability rises further, to approximately 96\% ($CI_{.95}$ [96\%, 97\%]) for no-reasoning agents and 98\% ($CI_{.95}$ [97\%, 98\%]) for high-reasoning agents, reversing the direction of the reasoning effect: high reasoning now \emph{increases} selection of the socially endorsed option. This indicates that LLM-based GUI agents were susceptible to both nudge types and that the reasoning configuration appears to moderate these effects in opposing directions. Adding the provider control (Model~2, Figure~\ref{fig:results_m2}) and the provider and size controls (Model~3, Figure~\ref{fig:results_m3}) leaves the estimated probabilities virtually unchanged, indicating that the condition and interaction effects are robust to differences between model families and scales.

Having read the estimated probabilities off Figures~\ref{fig:results_m1}---\ref{fig:results_m3}, we now turn to the coefficient-level estimates that formally test our hypotheses.
Table~\ref{tab:regression} presents the results of the three pooled Bayesian multilevel logistic regressions. Model (1) serves as the primary inferential model, while Models (2) and (3) add the provider and size controls. All condition and interaction coefficients are essentially identical across the three specifications, so we report the estimates of Model (1) in the text. Complete posterior summaries for all three models, including credible intervals and odds ratios, are provided in Appendix~\ref{app:pooled}, and the corresponding per-provider and per-model robustness fits are reported in Appendices~\ref{app:provider} and \ref{app:permodel}.

The agent random-intercept standard deviation ($1.15$, $1.11$, and $1.10$ across the three specifications) quantifies how much agents differ, on the log-odds scale, in their baseline propensity to select the target once condition and reasoning are accounted for. Behaviourally, this variance reflects the within-run consistency of each agent's six choices together with the residual stochasticity that sampling introduces, rather than durable participant-level heterogeneity.

Starting with H1, the pooled models in Table~\ref{tab:regression} reveal credible positive effects for both the default nudge condition ($\beta = 2.48$, $CI_{.95}$ $[2.28,\, 2.68]$) and the social influence nudge condition ($\beta = 3.32$, $CI_{.95}$ $[3.10,\, 3.55]$), indicating that both nudge types substantially increased the likelihood of choosing the target product relative to the no-nudge baseline. 
Expressed as odds ratios, the default nudge multiplied the odds of selecting the target product by a factor of $8.70$ ($CI_{.95}$ $[7.54,\, 9.97]$) and the social influence nudge by a factor of $37.24$ ($CI_{.95}$ $[31.04,\, 44.20]$; Table~\ref{tab:app-contrasts}). Both contrasts remain virtually unchanged under the provider and size controls ($OR$ =  $8.94$ for defaults, and $38.69$ for social influence).
These effects are corroborated by all six per-model regressions, in which both the default and the social influence coefficients are credible for every model (Appendix~\ref{app:permodel}, Tables~D1--D6). In sum, these results provide robust support for H1. LLM-based GUI agents exhibit systematic shifts in choice behaviour when exposed to both automatic and reflective nudging interventions, regardless of the underlying model, provider and size.

For H2a, we tested whether agents under a high-reasoning configuration were less susceptible to automatic Type 1 nudges, that is, defaults. The pooled model shows that reasoning alone did not credibly affect behaviour in the no-nudge condition ($\beta = -0.05$, $CI_{.95}$ $[-0.21,\, 0.12]$, including zero), confirming that differences emerge only when a nudge is present. Critically, the interaction between the reasoning configuration and the default nudge condition is credibly negative ($\beta = -0.63$, $CI_{.95}$ $[-0.89,\, -0.36]$) and robust to the provider and size controls ($\beta = -0.62$, $CI_{.95}$ $[-0.88,\, -0.36]$), indicating that high-reasoning agents were less affected by default nudges than no-reasoning agents. 

\begin{table}[ht]
\caption{Bayesian multilevel logistic regression results for the three pooled models. Model (1) regresses target selection on condition, reasoning, and their interaction. Model (2) adds the provider control, and Model (3) adds the provider and size controls. Cell entries are posterior means (log-odds) with posterior standard deviations in parentheses. Complete posterior summaries, including 95\% credible intervals and odds ratios, are reported in Appendix~\ref{app:pooled}.}\label{tab:regression}
\begin{tabular*}{\textwidth}{@{\extracolsep\fill}lccc}
\toprule
~ & \multicolumn{3}{c}{\textbf{Target Selected}} \\
\midrule
 & (1) & (2) & (3) \\
\textbf{Dependent Variable:} & Pooled & + Provider & + Provider + Size \\
\midrule
\midrule
Intercept & $-0.02$ & $0.54$* & $0.46$* \\
          & $(0.06)$ & $(0.07)$ & $(0.08)$ \\[3pt]
Defaults & $2.48$* & $2.50$* & $2.50$* \\
                   & $(0.10)$ & $(0.10)$ & $(0.10)$ \\[3pt]
Social Influence & $3.32$* & $3.37$* & $3.35$* \\
                           & $(0.11)$ & $(0.11)$ & $(0.11)$ \\[3pt]
Reasoning high & $-0.05$ & $-0.05$ & $-0.05$ \\
               & $(0.08)$ & $(0.08)$ & $(0.08)$ \\[3pt]
Provider Google\textsuperscript{\textbf{a}} & -- & $-0.99$* & $-0.98$* \\
                & -- & $(0.07)$ & $(0.07)$ \\[3pt]
Provider OpenAI\textsuperscript{\textbf{a}} & -- & $-0.68$* & $-0.69$* \\
                & -- & $(0.07)$ & $(0.07)$ \\[3pt]
Size small\textsuperscript{\textbf{b}} & -- & -- & $0.17$* \\
           & -- & -- & $(0.06)$ \\[3pt]
Defaults $\times$ Reas.\ high & $-0.63$* & $-0.62$* & $-0.62$* \\
                                     & $(0.13)$ & $(0.13)$ & $(0.13)$ \\[3pt]
Soc.\ Infl.\ $\times$ Reas.\ high & $0.59$* & $0.60$* & $0.60$* \\
                                         & $(0.16)$ & $(0.16)$ & $(0.16)$ \\[3pt]
\midrule
\multicolumn{4}{c}{* 95\% credible interval excludes zero} \\
\midrule
 & (1) & (2) & (3) \\
\midrule
Std.\ dev.\ agent intercept & $1.15$ & $1.11$ & $1.10$ \\
Observations (agents) & $21{,}600\ (3{,}600)$ & $21{,}600\ (3{,}600)$ & $21{,}600\ (3{,}600)$ \\
\botrule
\end{tabular*}
\footnotetext{Priors are weakly informative Student-$t(3, 0, 2.5)$ on all coefficients, with 4 chains of 4\,000 iterations (1\,000 warmup) each. All $\widehat{R} \approx 1.00$.}
\footnotetext{\textsuperscript{\textbf{a}}Provider controls (Anthropic as the reference category), included to verify that the condition and interaction effects are robust to model family. Provider is treated purely as a control.}
\footnotetext{\textsuperscript{\textbf{b}}Model size control (flagship as the reference category). The size coefficient is interpreted as an index of model scale (see Section~\ref{sec:results}).}
\end{table}

In odds-ratio terms, within the default condition, high reasoning roughly halved the odds of selecting the preselected product ($OR = 0.51$, $CI_{.95}$ $[0.41,\, 0.62]$; Table~\ref{tab:app-contrasts}). This is also visible in Figure~\ref{fig:results_m1}, where the predicted probability drops from approximately 92\% for no-reasoning agents to approximately 86\% for high-reasoning agents. The results thus support H2a.

For H2b, we examined whether agents under a high-reasoning configuration were more susceptible to reflective Type 2 nudges, that is, social influence. The pooled model shows that the interaction between the reasoning configuration and the social influence condition is credibly positive ($\beta = 0.59$, $CI_{.95}$ $[0.27,\, 0.91]$) and, like the H2a interaction, robust to the provider and size controls ($\beta = 0.60$, $CI_{.95}$ $[0.28,\, 0.92]$), indicating that the social influence effect was on average stronger for high-reasoning agents. In odds-ratio terms, within the social influence condition, high reasoning multiplied the odds of selecting the socially endorsed product by $1.71$ ($CI_{.95}$ $[1.26,\, 2.20]$; Table~\ref{tab:app-contrasts}), corresponding to an increase in the predicted probability from approximately 96\% to 98\%. The results thus support H2b.

\FloatBarrier
\subsection{Post hoc Exploratory Analysis}\label{subsec:res-exp}

The pooled models estimate the reasoning moderation as an average across six models that differ in scale, architecture, training, and alignment. Such an average can obscure systematic heterogeneity between model families and scales \citep{wei2022emergent,mckenzie2023inverse,perez2023discovering}. We therefore examine, in an exploratory analysis not anticipated by our hypotheses, whether the reasoning moderation is uniform across models or varies systematically with model scale and provider. To this end, we re-estimate Model~1 specification separately for each  provider,  each size class, and each individual model (Appendices~\ref{app:provider}--\ref{app:permodel}).

At the individual-model level, the two effects of reasoning appear in different models. High reasoning weakens the default nudge mainly in the larger flagship models. In Gemini 3.5 Flash (Table \ref{tab:gemini-3-5-flash}), for instance, it lowers the share of agents that keep the preselected default from 84\% to 48\%, essentially chance ($\beta = -1.61$, $CI_{.95}$ $[-2.30,\, -0.96]$), whereas in its smaller counterpart Gemini 3.1 Flash Lite (Table \ref{tab:gemini-3-1-flash-lite}), the change is less. High reasoning instead strengthens the social influence nudge mainly in the smaller models. In GPT-5.4-mini (Table \ref{tab:GPT-5.4-mini}), for instance, it raises the share that follow the socially endorsed option from 91\% to 99\% ($\beta = 2.46$, $CI_{.95}$ $[1.61,\, 3.38]$), whereas its larger counterpart GPT-5.4 (Table \ref{tab:GPT-5.4}) shows no such increase.

Pooling the models by size class confirms this reasoning-based split at the aggregate level (Appendix~\ref{app:size}) and surfaces a second, opposing pattern in overall susceptibility. Independently of reasoning, the default nudge pulls the smaller variants more strongly than the flagships ($OR = 15.94$ versus $OR = 4.64$; Table \ref{Posterior_size}), whereas the social influence nudge pulls the flagships more strongly ($OR = 107.57$ versus $OR = 17.59$; Table \ref{Posterior_size}). Overall susceptibility therefore runs opposite to the reasoning moderation, with defaults dominating in the smaller models and social influence in the flagships. The per-provider fits are consistent with these patterns (Appendix~\ref{app:provider}).

A caveat concerns the two Claude models, which reach the 100\% ceiling in entire conditions (Claude Sonnet 4.6 under social influence at both reasoning configurations, and Claude Haiku 4.5 under defaults). This (quasi-)complete separation renders the affected point estimates partly prior-dependent, so the corresponding conclusions rest on the data-driven credible interval bounds rather than the point estimates. The complete per-model analyses are reported in Appendix~\ref{app:permodel}.

Taken together, these breakdowns point to a structure that tracks model scale. Larger flagship models account for most of the reasoning-driven weakening of the default nudge, while the smaller variants account for most of the reasoning-driven strengthening of the social influence nudge. The raw pull of each nudge, however, follows the reverse ordering, since defaults weigh more heavily on the small models and social cues on the flagships. As these regularities were  hypothesized, we read them as tentative and leave their confirmation to future work.

\section{Discussion, Limitations and Outlook}\label{sec:discussion}

This study investigates how LLM-based GUI agents behave when the digital environments they navigate are designed to influence their decisions. Drawing on Dual-Process Theory, we examined whether behavioural interventions originally developed for humans---automatic (Type 1) and reflective (Type 2) nudges---affect AI agents, and whether the reasoning configuration of an agent moderates its susceptibility. Across six leading models from three providers, including both flagship and smaller variants, we provide broad empirical evidence and insights into how LLM-based GUI agents succumb to digital nudges.

Our findings demonstrate that LLM-based agents are highly susceptible to behavioural nudges, thereby challenging the assumption that they act as purely rational decision-makers \citep{chen2023emergence}. Interestingly, in our online shopping task, agents were even more responsive to these nudges than the human participants in the experiment on which our study was based \citep{ingendahlWhoCanBe2021}. 
In sum, a preselected default increased the likelihood of selecting the nudged product roughly ninefold, while indicating that other shoppers had preferred it raised the odds more than thirtyfold, a susceptibility observed in every one of the six models. Our findings therefore extend and add to previous evidence of human-like cognitive biases in LLMs from text-based settings \citep{nguyen2024human,hagendorff2023human,cherep2024superficial} to agentic environments, demonstrating that such  biases persist when models operate as autonomous agents that perceive interfaces, plan sequences of actions, and execute decisions.

The effects of additional reasoning reveal a more complex relationship between deliberation and robustness. Extended reasoning reduced susceptibility to the default nudge by approximately half, consistent with dual-process accounts in which reflective processing can override automatic responses \citep{stanovichRationalityReflectiveMind2011}. However, the same reasoning configuration increased susceptibility to the social influence nudge by approximately 70\%. Thus, deliberation did not eliminate behavioural bias in LLM agents.  Instead, it changed the type of influence to which agents were most responsive. While reflective processing appears to reduce reliance on passive defaults, it may increase attention to socially relevant information that appears informative or normatively appropriate.

This opposing pattern is consistent with a dual-process interpretation. Because System 2-style reasoning reduces the intuitive errors associated with automatic processing \citep{hagendorff2023human,zhang2025system}, a high reasoning configuration may allow agents to override the automatic pull of a preselected option. Reflective nudges, by contrast, take effect when their informational content is processed. More extensive reasoning may therefore amplify attention to, and deepen engagement with the social cues, similar to how humans consciously weigh and incorporate perceived norms when forming judgments \citep{cialdiniSocialInfluenceSocial1998,deutschStudyNormativeInformational1955}. This finding also aligns with research showing that models trained with human feedback increasingly conform to stated views and social signals, a tendency documented as sycophancy \citep{perez2023discovering,sharma2024towards} and majority conformity \citep{weng2025we}. Because reasoning-oriented training also emphasizes instruction-following and value alignment \citep{zhang2025system,lynchAgenticMisalignmentHow2025}, agents configured for more deliberation may be especially responsive to persuasive informational cues.

Our exploratory analysis further suggests that these effects may vary by model scale. Reasoning-based resistance to defaults was primarily observed in flagship models, potentially because overriding a default requires generating and acting upon an alternative preference—a capability that may improve with greater model capacity \citep{wei2022emergent}. Conversely, increased susceptibility to social influence was more pronounced among smaller variants. One plausible explanation is that the flagship models were already highly compliant with the social cue under the no-reasoning configuration, leaving little room for further increases, whereas the smaller variants retained greater scope for deeper processing of the cue to increase compliance. We offer this interpretation cautiously, as our study was not designed to test this mechanism directly. Accordingly, we treat these findings as exploratory and as a basis for future confirmatory research.

Our findings also contribute to understanding the relationship between agentic and human decision-making. In the human study adapted here, \citet{ingendahlWhoCanBe2021} found both nudges to be effective. However, need for cognition—a tendency to engage in effortful, reflective thinking \citep{cacioppo1982need}—only weakly and inconsistently moderated these effects, with less reflective participants showing slightly greater susceptibility to the default nudge.
While the role of cognitive engagement in human nudgeability more generally remains debated \citep{deridderNudgeabilityMappingConditions2022,vangestelNudgesMakeUse2021}, directly manipulating the reasoning configuration in agents produced a stronger and more systematic moderation effect. 
Agents under high reasoning were markedly less susceptible to the automatic default nudge, in the same direction as the human default tendency, while becoming more susceptible to the reflective social influence nudge. We draw this comparison cautiously, since switching an agent's reasoning on or off is not the same as comparing people who naturally differ in a trait such as need for cognition. It nevertheless suggests that agent-based simulations can offer a complementary setting for examining how processing mode shapes susceptibility to nudges. 
Notably, these effects emerged in the absence of the cognitive constraints traditionally invoked to explain human bias \citep{liederResourcerationalAnalysisUnderstanding2020}, suggesting that agentic susceptibility may arise from mechanisms such as training data, learned heuristics, and alignment objectives \citep{jones2022capturing,macmillan2024ir}. Susceptibility to reflective nudges thus appears to be not merely a consequence of limited cognition but a byproduct of how both humans and agents use social information as a shortcut for what is appropriate or accurate \citep{caraban23WaysNudge2019}.

From a theoretical perspective, our results establish nudge susceptibility as a behavioural property of agentic AI systems rather than a phenomenon unique to humans \citep{bairdNextGenerationResearch2021,bosch2025biased,cherep2024superficial}. Because the reflective-moderates-automatic pattern appears in agents that lack the cognitive architecture and resource limits Dual-Process Theory assumes \citep{kahneman2003maps,evansDualProcessTheoriesHigher2013}, the dual-process regularity is separable from its human substrate and may characterise any instruction-following, preference-aligned decision system—extending the theory beyond human cognition rather than merely applying it. That the pattern recurs across model families differing in architecture, training data, and alignment points to a general property of current LLM-based agents rather than an artifact of any single model. The manipulation also reframes deliberation from debiasing to redirection: unlike the human literature, where cognitive engagement is a contested trait moderator \citep{vangestelNudgesMakeUse2021,deridderNudgeabilityMappingConditions2022}, our experimental control isolates an opposing-direction effect that qualifies the expectation that more reasoning yields more robust decisions \citep{zhang2025system,mckenzie2023inverse}. The scale-dependence we observe further suggests this moderation is itself capacity-bounded, a condition with no direct analogue in the human formulation.
Taken together, these points underscore the value of evaluating agentic systems behaviourally, looking beyond task competence toward how context and environment shape their decisions and can misalign them from their intended behaviour \citep{lynchAgenticMisalignmentHow2025}.

From a practical perspective, our results position interface design as a governance concern rather than a matter of usability alone. The nudges studied here are ordinary features of digital choice architectures \citep{schneiderDigitalNudgingGuiding2018,caraban23WaysNudge2019}, not adversarial attacks such as prompt injection \citep{greshake2023not}. Agents are therefore routinely exposed to them, and because digital choice architectures shape agent decisions much as they shape human ones, the consequences of these design choices ultimately accrue to the delegating user. Deploying agents responsibly thus requires behavioural auditing, evaluation across diverse interface conditions, and safeguards against unintended influence.

These considerations may become even more important as AI agents grow increasingly capable. Although advanced agents require less step-by-step prompting and can operate more independently,\footnote{\url{https://www.anthropic.com/engineering/building-effective-agents}} they may remain susceptible to subtle contextual cues---a vulnerability that could persist even as other forms of bias diminish, precisely because agents are trained to follow instructions and align with human preferences. Cues that appear consistent with those preferences may therefore steer behaviour towards unintended outcomes \citep{zhang2025system}.

These risks call for safeguards beyond conventional alignment techniques, distributed across the actors who build, deploy, and host agents. Developers could report standardized nudge-susceptibility benchmarks that score a model with specific reasoning configuration and scale against common interface manipulation. Organizations that deploy agents could treat the reasoning configuration as a deployment decision rather than a default, choosing it for the specific choice environment while logging the agent's choice distribution to detect behavioural drift when an interface changes, and requiring a verification pass that re-checks consequential actions against the task's stated criteria before they are executed. Interface and platform designers, in turn, could support machine-readable disclosure of choice-architecture elements, such as tagging a preselected default or a social-proof badge, so that a delegated agent can recognise and discount a persuasive cue it would otherwise treat as neutral information. More broadly, our findings point to a capability that cuts across these roles and remains difficult to achieve. Agents would need to tell information relevant to the task apart from persuasive cues that are not, and to give less weight to the latter. Our results suggest this is unlikely to come from more reasoning alone, which for reflective nudges can even increase susceptibility. Together, such measures could improve the robustness of agentic systems to routine interface manipulations without compromising their ability to interact effectively with digital environments, and they motivate future work on enabling agents to recognise, interpret, and appropriately respond to contextual cues without being unduly influenced by them.

As with any study, this work must be considered in light of some limitations. First, we examined only default and social influence nudges as representative mechanisms of Type 1 and Type 2 influence. Whether the same pattern extends to other automatic nudges, such as deceptive visualizations or scarcity cues, and to other reflective nudges, such as suggesting alternatives or reminding of consequences, remains open. Second, we manipulated reasoning through the model configuration exposed by the providers' APIs. Future work should examine complementary strategies, such as prompt-level instructions that encourage deliberation about product attributes or trade-offs, to distinguish task-specific reasoning from the general reasoning configuration, and should test how agents respond when several nudges are present at once. Third, the scale-dependent structure of the moderation effects was not hypothesized and is reported as exploratory. Confirming it will require designs with more models per scale class, so that model size can be treated as a planned factor rather than inferred post hoc.
Fourth, our findings characterise the specific model versions used in this study, whose exact configurations are documented in our public repository. Because providers update weights, alignment, and API behaviour under stable model names, the absolute susceptibility levels---and potentially the reasoning moderation---may differ for later versions, even where the model name is unchanged. The consistency of the central pattern across six models from three providers nonetheless makes it unlikely to be an artifact of any single version. 
Finally, regarding external validity, our experiment adapted an established human decision-making experiment within a simulated online store. While this controlled setting offers experimental rigour, future research should study LLM-based GUI agents in richer and more ecologically valid environments to determine whether the observed effects persist beyond laboratory conditions. Deploying such agents safely will depend less on making them more capable than on scrutinising the environments in which they are asked to decide.

\section{Conclusion}\label{sec:conclusion}

With growing delegation of tasks to autonomous AI agents across domains such as e-commerce and finance, understanding how they behave in environments designed to influence them becomes increasingly important.
In this paper, we examined one such form of external influence, asking how digital nudges shape the choice behaviour of LLM-based GUI agents. By adapting a controlled human decision-making experiment to six leading models from three providers, we showed that agents are susceptible to both automatic and reflective nudges, and that reasoning does not immunise them but changes which nudges take hold---weakening automatic defaults while strengthening reflective social cues. An exploratory analysis indicated that this pattern is systematically structured by model size.

\backmatter

\bibliography{references_manuell}

\begin{appendices}

\counterwithin{table}{section}
\counterwithin{figure}{section}
\renewcommand{\thetable}{\thesection\arabic{table}}
\renewcommand{\thefigure}{\thesection\arabic{figure}}

\section{Complete posterior summaries for the pooled models}\label{app:pooled}

Tables~\ref{tab:app-m1}--\ref{tab:app-m3} report the complete posterior summaries of the three pooled models presented in Table~\ref{tab:regression}, namely posterior means and posterior standard deviations (Est.\ Error) with 95\% credible intervals on the log-odds scale, together with the corresponding odds ratios ($OR = e^{\beta}$) and their 95\% credible intervals. Table~\ref{tab:app-contrasts} additionally reports the posterior odds-ratio contrasts underlying the hypothesis tests, namely the pairwise condition contrasts (H1) and the high versus no reasoning contrasts within each condition (H2a and H2b), summarized by posterior medians with 95\% highest-posterior-density intervals.

\begin{table}[ht]
\caption{Model (1), pooled, complete posterior summary ($N = 21{,}600$ observations, $3{,}600$ agents).}\label{tab:app-m1}
\begin{tabular*}{\textwidth}{@{\extracolsep\fill}lrrrrrrr}
\toprule
~ & ~ & ~ & \multicolumn{2}{c}{95\% CI} & ~ & \multicolumn{2}{c}{OR 95\% CI} \\
\cmidrule{4-5}\cmidrule{7-8}
Parameter & Estimate & Est.\ Error & lower & upper & OR & lower & upper \\
\midrule
Intercept & $-0.02$ & $0.06$ & $-0.13$ & $0.10$ & $0.98$ & $0.87$ & $1.10$ \\
Defaults & $2.48$ & $0.10$ & $2.28$ & $2.68$ & $11.94$ & $9.80$ & $14.56$ \\
Social Influence & $3.32$ & $0.11$ & $3.10$ & $3.55$ & $27.78$ & $22.28$ & $34.70$ \\
Reasoning high & $-0.05$ & $0.08$ & $-0.21$ & $0.12$ & $0.95$ & $0.81$ & $1.12$ \\
Defaults $\times$ Reas.\ high & $-0.63$ & $0.13$ & $-0.89$ & $-0.36$ & $0.53$ & $0.41$ & $0.69$ \\
Soc.\ Infl.\ $\times$ Reas.\ high & $0.59$ & $0.16$ & $0.27$ & $0.91$ & $1.80$ & $1.31$ & $2.48$ \\
\midrule
Std.\ dev.\ agent intercept & $1.15$ & $0.04$ & $1.08$ & $1.22$ & -- & -- & -- \\
\botrule
\end{tabular*}
\end{table}

\begin{table}[ht]
\caption{Model (2), pooled + provider control, complete posterior summary ($N = 21{,}600$ observations, $3{,}600$ agents).}\label{tab:app-m2}
\begin{tabular*}{\textwidth}{@{\extracolsep\fill}lrrrrrrr}
\toprule
~ & ~ & ~ & \multicolumn{2}{c}{95\% CI} & ~ & \multicolumn{2}{c}{OR 95\% CI} \\
\cmidrule{4-5}\cmidrule{7-8}
Parameter & Estimate & Est.\ Error & lower & upper & OR & lower & upper \\
\midrule
Intercept & $0.54$ & $0.07$ & $0.40$ & $0.68$ & $1.71$ & $1.49$ & $1.97$ \\
Defaults & $2.50$ & $0.10$ & $2.31$ & $2.70$ & $12.18$ & $10.03$ & $14.93$ \\
Social Influence & $3.37$ & $0.11$ & $3.15$ & $3.59$ & $29.02$ & $23.42$ & $36.22$ \\
Reasoning high & $-0.05$ & $0.08$ & $-0.21$ & $0.11$ & $0.95$ & $0.81$ & $1.12$ \\
Provider Google & $-0.99$ & $0.07$ & $-1.13$ & $-0.84$ & $0.37$ & $0.32$ & $0.43$ \\
Provider OpenAI & $-0.68$ & $0.07$ & $-0.83$ & $-0.54$ & $0.50$ & $0.44$ & $0.58$ \\
Defaults $\times$ Reas.\ high & $-0.62$ & $0.13$ & $-0.88$ & $-0.37$ & $0.54$ & $0.41$ & $0.69$ \\
Soc.\ Infl.\ $\times$ Reas.\ high & $0.60$ & $0.16$ & $0.28$ & $0.92$ & $1.82$ & $1.33$ & $2.51$ \\
\midrule
Std.\ dev.\ agent intercept & $1.11$ & $0.04$ & $1.04$ & $1.18$ & -- & -- & -- \\
\botrule
\end{tabular*}
\end{table}

\begin{table}[ht]
\caption{Model (3), pooled + provider + size controls, complete posterior summary ($N = 21{,}600$ observations, $3{,}600$ agents).}\label{tab:app-m3}
\begin{tabular*}{\textwidth}{@{\extracolsep\fill}lrrrrrrr}
\toprule
~ & ~ & ~ & \multicolumn{2}{c}{95\% CI} & ~ & \multicolumn{2}{c}{OR 95\% CI} \\
\cmidrule{4-5}\cmidrule{7-8}
Parameter & Estimate & Est.\ Error & lower & upper & OR & lower & upper \\
\midrule
Intercept & $0.46$ & $0.08$ & $0.30$ & $0.61$ & $1.58$ & $1.35$ & $1.84$ \\
Defaults & $2.50$ & $0.10$ & $2.30$ & $2.70$ & $12.19$ & $10.02$ & $14.91$ \\
Social Influence & $3.35$ & $0.11$ & $3.14$ & $3.58$ & $28.63$ & $23.03$ & $35.86$ \\
Reasoning high & $-0.05$ & $0.08$ & $-0.21$ & $0.11$ & $0.95$ & $0.81$ & $1.12$ \\
Provider Google & $-0.98$ & $0.07$ & $-1.13$ & $-0.84$ & $0.37$ & $0.32$ & $0.43$ \\
Provider OpenAI & $-0.69$ & $0.07$ & $-0.83$ & $-0.54$ & $0.50$ & $0.43$ & $0.58$ \\
Size small & $0.17$ & $0.06$ & $0.05$ & $0.28$ & $1.18$ & $1.06$ & $1.32$ \\
Defaults $\times$ Reas.\ high & $-0.62$ & $0.13$ & $-0.88$ & $-0.36$ & $0.54$ & $0.42$ & $0.70$ \\
Soc.\ Infl.\ $\times$ Reas.\ high & $0.60$ & $0.16$ & $0.29$ & $0.92$ & $1.83$ & $1.33$ & $2.51$ \\
\midrule
Std.\ dev.\ agent intercept & $1.10$ & $0.04$ & $1.03$ & $1.17$ & -- & -- & -- \\
\botrule
\end{tabular*}
\end{table}

\begin{table}[ht]
\caption{Posterior odds-ratio contrasts for the pooled sample, comprising condition contrasts (H1) and high versus no reasoning contrasts within each condition (H2a and H2b). Entries are posterior medians with 95\% highest-posterior-density intervals in brackets.}\label{tab:app-contrasts}
\small
\begin{tabular*}{\textwidth}{@{\extracolsep{\fill}}lccc}
\toprule
Contrast & (1) & (2) & (3) \\
         & Pooled & \makecell{+ Provider} & \makecell{+ Provider\\+ Size} \\
\midrule
Defaults / No nudge
  & $8.70$          & $8.92$           & $8.94$ \\
  & {\footnotesize $[7.54,\,9.97]$}   & {\footnotesize $[7.73,\,10.17]$}  & {\footnotesize $[7.76,\,10.23]$} \\[2pt]
Social influence / No nudge
  & $37.24$         & $39.16$          & $38.69$ \\
  & {\footnotesize $[31.04,\,44.20]$} & {\footnotesize $[32.95,\,46.51]$} & {\footnotesize $[32.58,\,45.81]$} \\[2pt]
Social influence / Defaults
  & $4.27$          & $4.39$           & $4.33$ \\
  & {\footnotesize $[3.58,\,5.05]$}   & {\footnotesize $[3.64,\,5.16]$}   & {\footnotesize $[3.63,\,5.11]$} \\
\midrule
High / no reasoning, No nudge
  & $0.95$          & $0.95$           & $0.95$ \\
  & {\footnotesize $[0.81,\,1.12]$}   & {\footnotesize $[0.81,\,1.12]$}   & {\footnotesize $[0.80,\,1.11]$} \\[2pt]
High / no reasoning, Defaults
  & $0.51$          & $0.51$           & $0.51$ \\
  & {\footnotesize $[0.41,\,0.62]$}   & {\footnotesize $[0.41,\,0.62]$}   & {\footnotesize $[0.41,\,0.62]$} \\[2pt]
High / no reasoning, Social influence
  & $1.71$          & $1.74$           & $1.74$ \\
  & {\footnotesize $[1.26,\,2.20]$}   & {\footnotesize $[1.30,\,2.27]$}   & {\footnotesize $[1.30,\,2.27]$} \\
\botrule
\end{tabular*}
\end{table}

\clearpage

\section{Per-provider pooled models}\label{app:provider}

\begin{figure}[ht]
\centering
\begin{subfigure}[b]{0.32\textwidth}
    \centering
    \includegraphics[width=\textwidth]{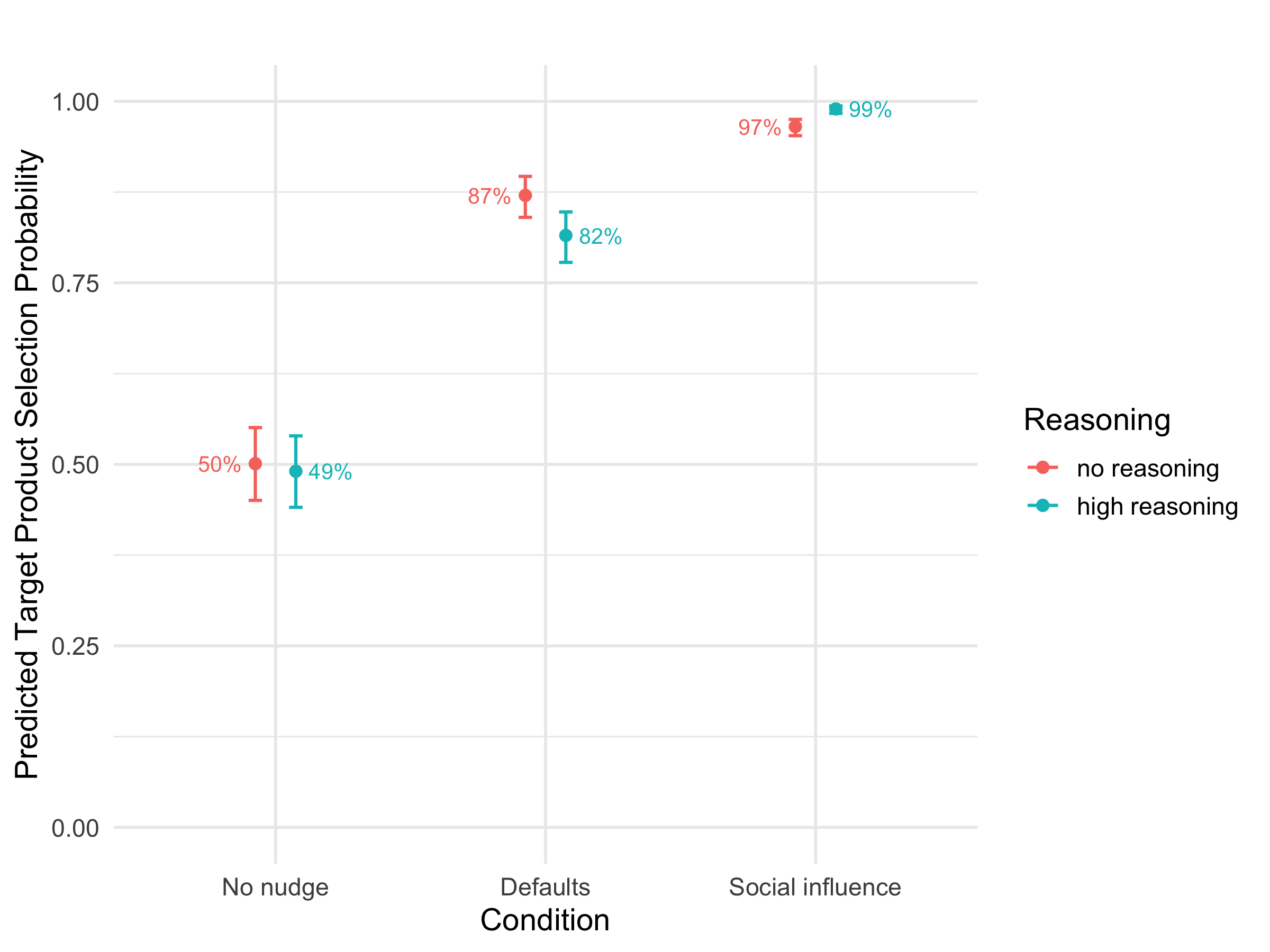}
    \caption{OpenAI}
    \label{fig:results_openai}
\end{subfigure}
\hfill
\begin{subfigure}[b]{0.32\textwidth}
    \centering
    \includegraphics[width=\textwidth]{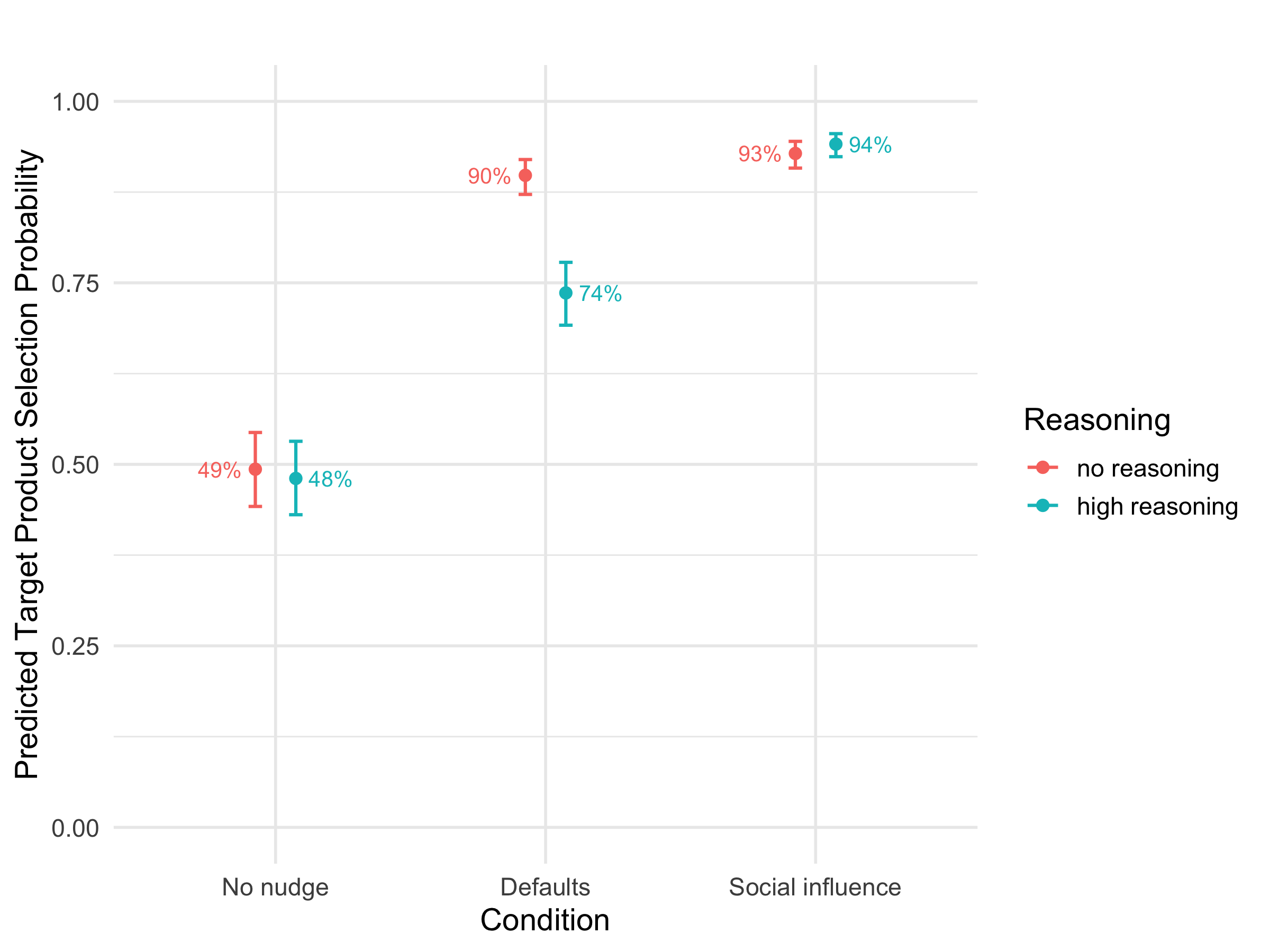}
    \caption{Google}
    \label{fig:results_google}
\end{subfigure}
\hfill
\begin{subfigure}[b]{0.32\textwidth}
    \centering
    \includegraphics[width=\textwidth]{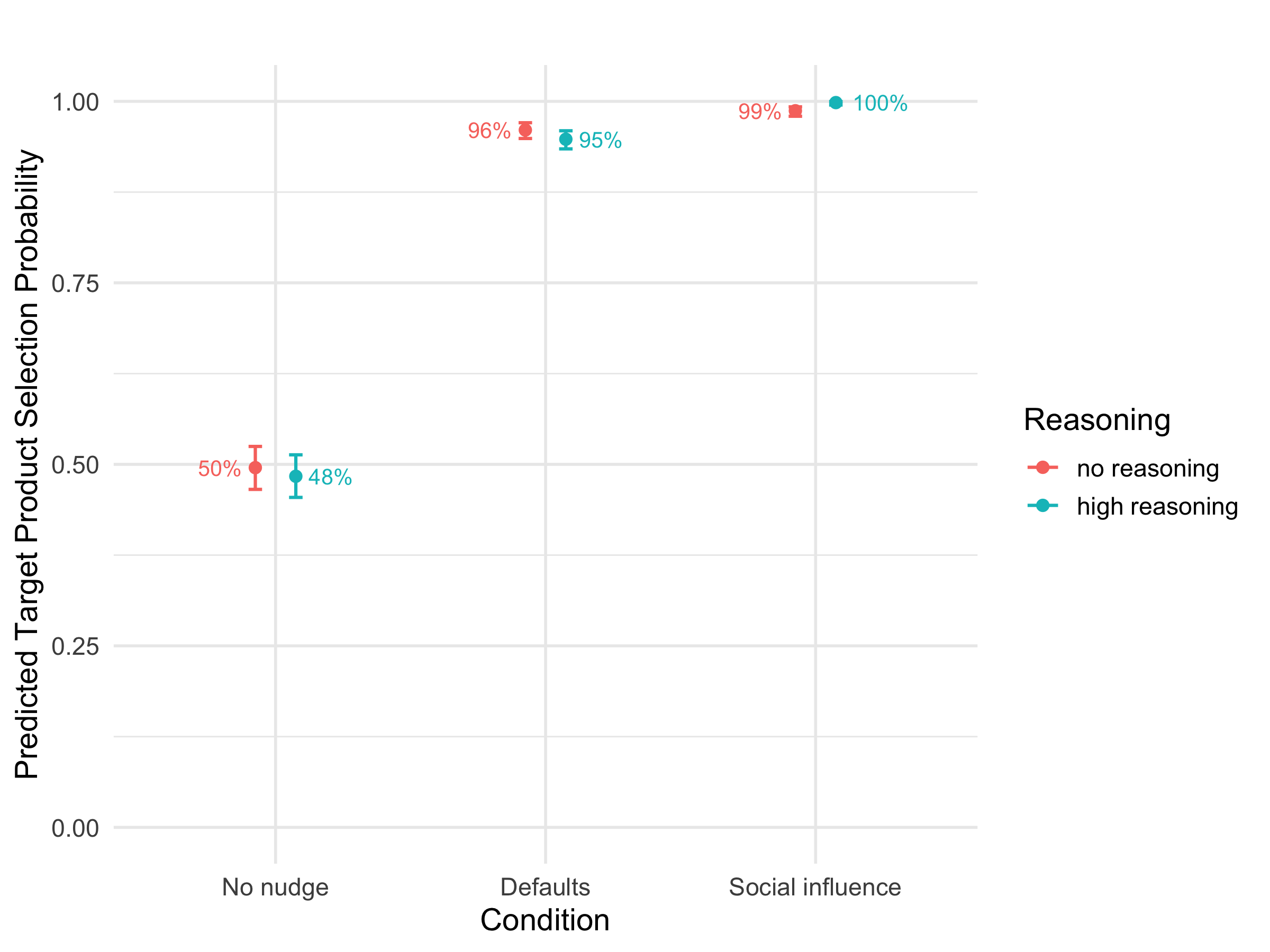}
    \caption{Anthropic}
    \label{fig:results_anthropic}
\end{subfigure}
\caption{Predicted target product selection probabilities across conditions and reasoning configurations for the per-provider pooled models. Red indicates no-reasoning agents, and blue indicates high-reasoning agents.}
\label{fig:results_provider}
\end{figure}

As a robustness check across providers, we re-estimated the Model (1) specification separately for each provider, pooling the provider's two models (flagship and small variant, $N = 7{,}200$ observations, $1{,}200$ agents each). Tables~\ref{tab:app-openai}--\ref{tab:app-anthropic} report the complete posterior summaries for the three provider pools, Table~\ref{tab:app-provider-contrasts} the corresponding odds-ratio contrasts (H1 and H2), and Figure~\ref{fig:results_provider} the predicted probabilities. The Claude ceiling cells apply to the Anthropic pool (see Appendix~\ref{app:permodel}). The results are consistent with the patterns discussed in Section~\ref{subsec:res-exp}.

\begin{table}[ht]
\caption{Provider pool OpenAI (GPT-5.4, GPT-5.4-mini), complete posterior summary.}\label{tab:app-openai}
\begin{tabular*}{\textwidth}{@{\extracolsep\fill}lrrrrrrr}
\toprule
~ & ~ & ~ & \multicolumn{2}{c}{95\% CI} & ~ & \multicolumn{2}{c}{OR 95\% CI} \\
\cmidrule{4-5}\cmidrule{7-8}
Parameter & Estimate & Est.\ Error & lower & upper & OR & lower & upper \\
\midrule
Intercept & $0.00$ & $0.10$ & $-0.20$ & $0.20$ & $1.00$ & $0.82$ & $1.23$ \\
Defaults & $1.90$ & $0.16$ & $1.59$ & $2.22$ & $6.70$ & $4.88$ & $9.24$ \\
Social Influence & $3.33$ & $0.20$ & $2.95$ & $3.73$ & $27.81$ & $19.05$ & $41.48$ \\
Reasoning high & $-0.04$ & $0.14$ & $-0.32$ & $0.24$ & $0.96$ & $0.73$ & $1.27$ \\
Defaults $\times$ Reas.\ high & $-0.38$ & $0.22$ & $-0.81$ & $0.05$ & $0.68$ & $0.44$ & $1.05$ \\
Soc.\ Infl.\ $\times$ Reas.\ high & $1.27$ & $0.32$ & $0.65$ & $1.90$ & $3.55$ & $1.92$ & $6.66$ \\
\midrule
Std.\ dev.\ agent intercept & $1.16$ & $0.06$ & $1.04$ & $1.28$ & -- & -- & -- \\
\botrule
\end{tabular*}
\end{table}

\begin{table}[ht]
\caption{Provider pool Google (Gemini 3.5 Flash, Gemini 3.1 Flash Lite), complete posterior summary.}\label{tab:app-google}
\begin{tabular*}{\textwidth}{@{\extracolsep\fill}lrrrrrrr}
\toprule
~ & ~ & ~ & \multicolumn{2}{c}{95\% CI} & ~ & \multicolumn{2}{c}{OR 95\% CI} \\
\cmidrule{4-5}\cmidrule{7-8}
Parameter & Estimate & Est.\ Error & lower & upper & OR & lower & upper \\
\midrule
Intercept & $-0.03$ & $0.10$ & $-0.23$ & $0.18$ & $0.97$ & $0.79$ & $1.19$ \\
Defaults & $2.20$ & $0.17$ & $1.87$ & $2.53$ & $9.05$ & $6.51$ & $12.54$ \\
Social Influence & $2.59$ & $0.18$ & $2.25$ & $2.94$ & $13.31$ & $9.45$ & $18.97$ \\
Reasoning high & $-0.05$ & $0.15$ & $-0.33$ & $0.24$ & $0.95$ & $0.72$ & $1.27$ \\
Defaults $\times$ Reas.\ high & $-1.10$ & $0.22$ & $-1.54$ & $-0.66$ & $0.33$ & $0.21$ & $0.52$ \\
Soc.\ Infl.\ $\times$ Reas.\ high & $0.26$ & $0.24$ & $-0.22$ & $0.75$ & $1.30$ & $0.80$ & $2.11$ \\
\midrule
Std.\ dev.\ agent intercept & $1.16$ & $0.06$ & $1.05$ & $1.28$ & -- & -- & -- \\
\botrule
\end{tabular*}
\end{table}

\begin{table}[ht]
\caption{Provider pool Anthropic (Claude Sonnet 4.6, Claude Haiku 4.5), complete posterior summary.}\label{tab:app-anthropic}
\begin{tabular*}{\textwidth}{@{\extracolsep\fill}lrrrrrrr}
\toprule
~ & ~ & ~ & \multicolumn{2}{c}{95\% CI} & ~ & \multicolumn{2}{c}{OR 95\% CI} \\
\cmidrule{4-5}\cmidrule{7-8}
Parameter & Estimate & Est.\ Error & lower & upper & OR & lower & upper \\
\midrule
Intercept & $-0.02$ & $0.06$ & $-0.14$ & $0.10$ & $0.98$ & $0.87$ & $1.10$ \\
Defaults & $3.21$ & $0.16$ & $2.91$ & $3.53$ & $24.84$ & $18.36$ & $34.28$ \\
Social Influence & $4.36$ & $0.27$ & $3.87$ & $4.91$ & $78.25$ & $48.15$ & $135.01$ \\
Reasoning high & $-0.05$ & $0.08$ & $-0.21$ & $0.12$ & $0.95$ & $0.81$ & $1.12$ \\
Defaults $\times$ Reas.\ high & $-0.24$ & $0.21$ & $-0.65$ & $0.17$ & $0.78$ & $0.52$ & $1.18$ \\
Soc.\ Infl.\ $\times$ Reas.\ high & $2.13$ & $0.73$ & $0.88$ & $3.76$ & $8.45$ & $2.40$ & $43.07$ \\
\midrule
Std.\ dev.\ agent intercept & $0.17$ & $0.11$ & $0.01$ & $0.39$ & -- & -- & -- \\
\botrule
\end{tabular*}
\end{table}

\begin{table}[ht]
\caption{Posterior odds-ratio contrasts for the three per-provider pooled models, comprising condition contrasts (H1) and high versus no reasoning contrasts within each condition (H2a and H2b). Entries are posterior medians with 95\% highest-posterior-density intervals. These are marginal contrasts (averaged over reasoning for H1), and therefore differ from the coefficient odds-ratios in Tables~\ref{tab:app-openai}--\ref{tab:app-anthropic}, which condition on the no-reasoning reference level.}\label{tab:app-provider-contrasts}
\begin{tabular*}{\textwidth}{@{\extracolsep\fill}lccc}
\toprule
Contrast & OpenAI pool & Google pool & Anthropic pool \\
\midrule
Defaults / No nudge & $5.54\ [4.38,\, 6.88]$ & $5.22\ [4.11,\, 6.51]$ & $21.92\ [17.52,\, 26.77]$ \\
Social influence / No nudge & $52.23\ [36.66,\, 72.25]$ & $15.14\ [11.68,\, 19.33]$ & $219.91\ [97.68,\, 449.90]$ \\
Social influence / Defaults & $9.42\ [6.55,\, 12.81]$ & $2.90\ [2.20,\, 3.68]$ & $10.02\ [4.26,\, 20.93]$ \\
\midrule
High / no reasoning, No nudge & $0.96\ [0.71,\, 1.24]$ & $0.95\ [0.70,\, 1.25]$ & $0.95\ [0.80,\, 1.12]$ \\
High / no reasoning, Defaults & $0.66\ [0.46,\, 0.90]$ & $0.32\ [0.22,\, 0.43]$ & $0.75\ [0.50,\, 1.07]$ \\
High / no reasoning, Social influence & $3.39\ [1.78,\, 5.64]$ & $1.24\ [0.79,\, 1.74]$ & $7.55\ [1.10,\, 30.21]$ \\
\botrule
\end{tabular*}
\end{table}

\clearpage

\section{Per-size pooled models}\label{app:size}

\begin{figure}[ht]
\centering
\begin{subfigure}[b]{0.45\textwidth}
    \centering
    \includegraphics[width=\textwidth]{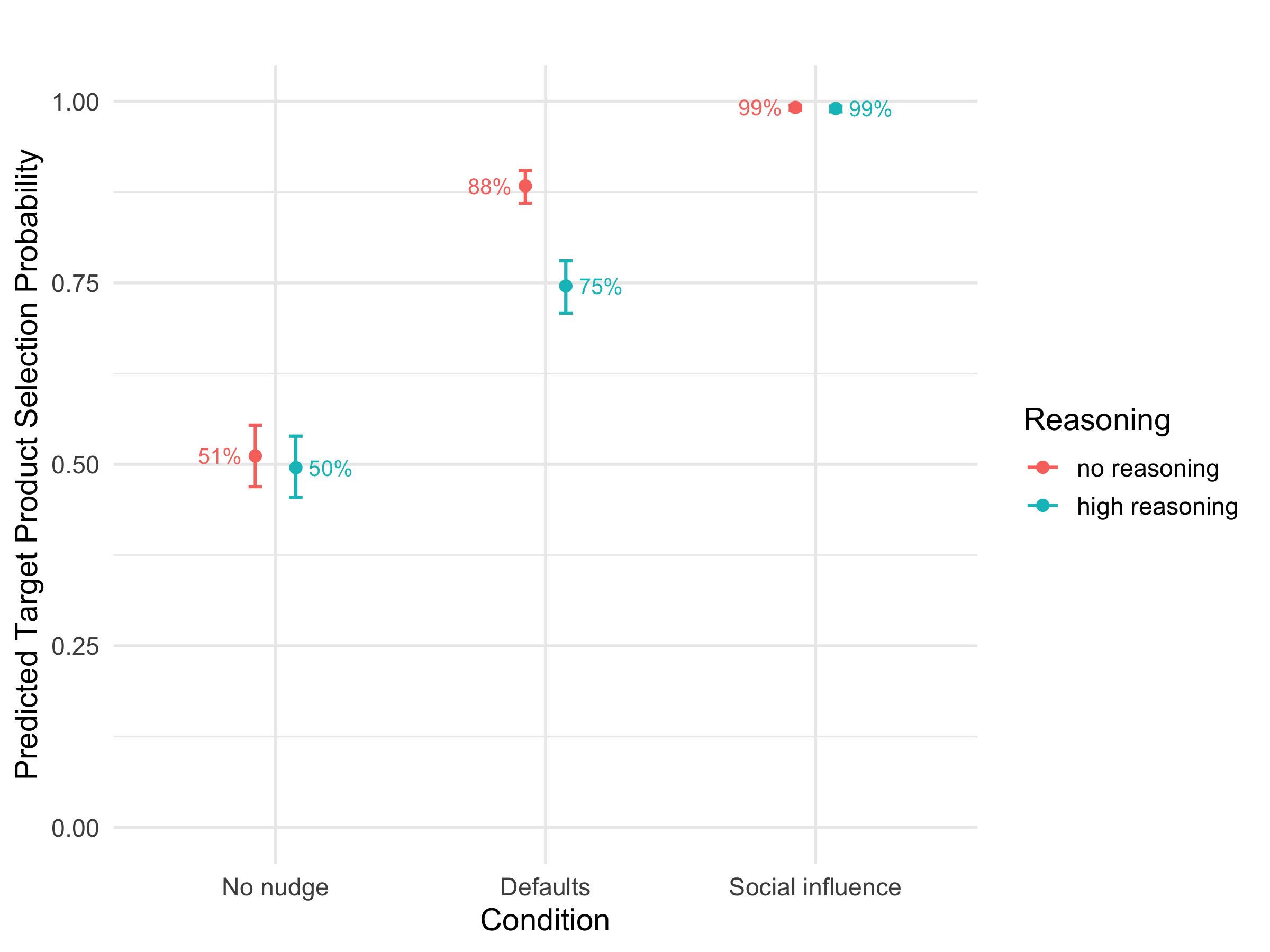}
    \caption{Flagship models}
    \label{fig:results_sizelarge}
\end{subfigure}
\hfill
\begin{subfigure}[b]{0.45\textwidth}
    \centering
    \includegraphics[width=\textwidth]{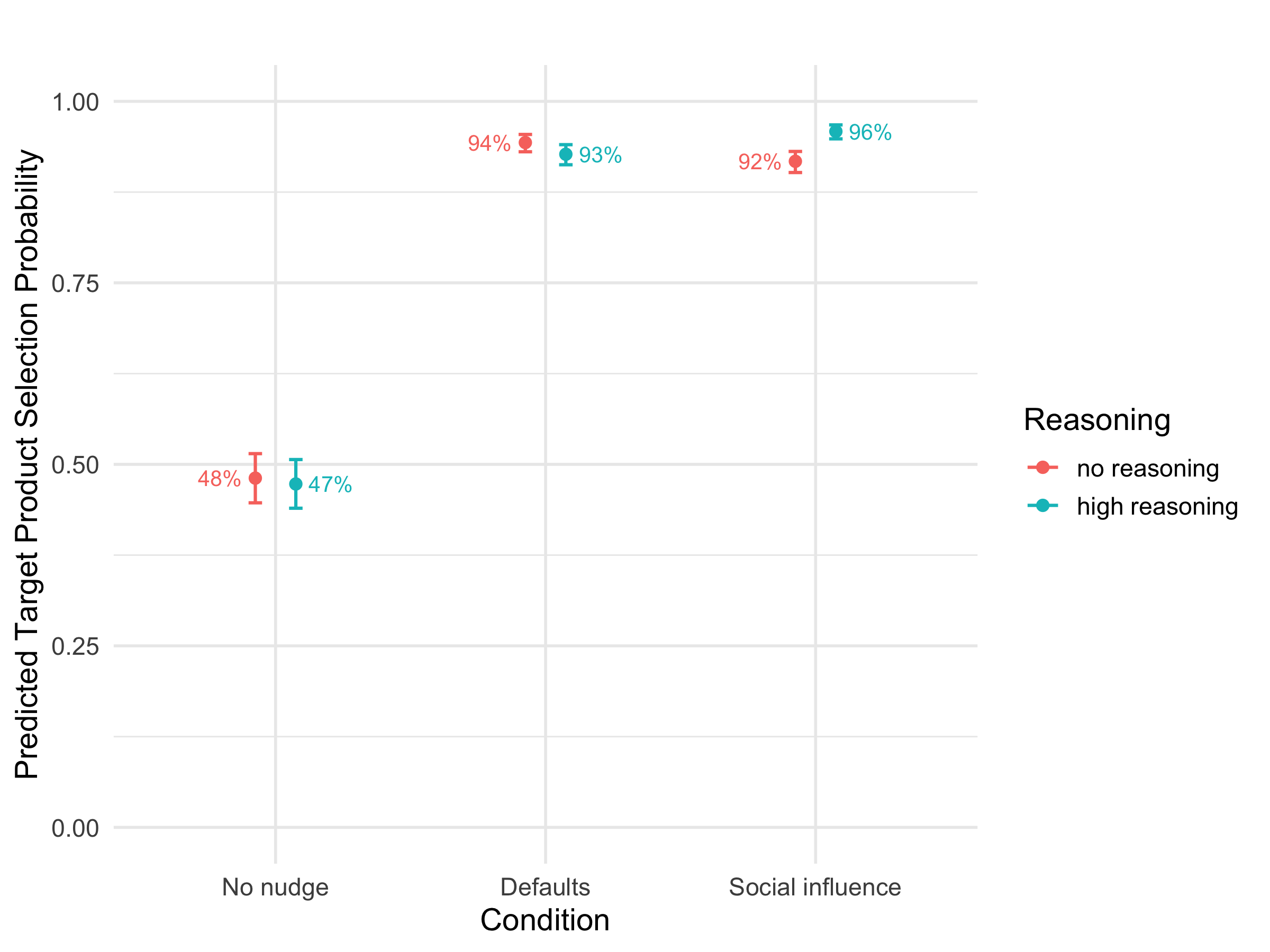}
    \caption{Small variants}
    \label{fig:results_sizesmall}
\end{subfigure}
\caption{Predicted target product selection probabilities across conditions and reasoning configurations for the per-size pooled models. Red indicates no-reasoning agents, and blue indicates high-reasoning agents.}
\label{fig:results_size}
\end{figure}

As an exploratory analysis of model scale, we re-estimated the Model (1) specification separately for each model size class, pooling together the three flagship models (GPT-5.4, Gemini 3.5 Flash, Claude Sonnet 4.6) and the three small variants (GPT-5.4-mini, Gemini 3.1 Flash Lite, Claude Haiku 4.5; $N = 10{,}800$ observations, $1{,}800$ agents each). Tables~\ref{tab:app-sizelarge} and \ref{tab:app-sizesmall} report the complete posterior summaries for the two size pools, Table~\ref{tab:app-size-contrasts} the corresponding odds-ratio contrasts (H1 and H2), and Figure~\ref{fig:results_size} the predicted probabilities. Two ceiling caveats apply: Claude Haiku 4.5's (quasi-)perfect default compliance contributes to the small pool's large default effect, and Claude Sonnet 4.6's perfect social influence compliance to the flagship pool's large social influence effect (see Appendix~\ref{app:permodel}). The results are consistent with the patterns discussed in Section~\ref{subsec:res-exp}.

\begin{table}[ht]
\caption{Size pool flagship (GPT-5.4, Gemini 3.5 Flash, Claude Sonnet 4.6), complete posterior summary.}\label{tab:app-sizelarge}
\begin{tabular*}{\textwidth}{@{\extracolsep\fill}lrrrrrrr}
\toprule
~ & ~ & ~ & \multicolumn{2}{c}{95\% CI} & ~ & \multicolumn{2}{c}{OR 95\% CI} \\
\cmidrule{4-5}\cmidrule{7-8}
Parameter & Estimate & Est.\ Error & lower & upper & OR & lower & upper \\
\midrule
Intercept & $0.05$ & $0.09$ & $-0.12$ & $0.22$ & $1.05$ & $0.88$ & $1.24$ \\
Defaults & $1.98$ & $0.14$ & $1.70$ & $2.26$ & $7.24$ & $5.50$ & $9.59$ \\
Social Influence & $4.73$ & $0.23$ & $4.29$ & $5.21$ & $113.52$ & $73.26$ & $182.87$ \\
Reasoning high & $-0.06$ & $0.12$ & $-0.31$ & $0.18$ & $0.94$ & $0.74$ & $1.20$ \\
Defaults $\times$ Reas.\ high & $-0.89$ & $0.19$ & $-1.25$ & $-0.52$ & $0.41$ & $0.29$ & $0.59$ \\
Soc.\ Infl.\ $\times$ Reas.\ high & $-0.10$ & $0.31$ & $-0.70$ & $0.49$ & $0.90$ & $0.49$ & $1.64$ \\
\midrule
Std.\ dev.\ agent intercept & $1.23$ & $0.05$ & $1.13$ & $1.34$ & -- & -- & -- \\
\botrule
\end{tabular*}
\end{table}

\begin{table}[ht]
\caption{Size pool small (GPT-5.4-mini, Gemini 3.1 Flash Lite, Claude Haiku 4.5), complete posterior summary.}\label{tab:app-sizesmall}
\begin{tabular*}{\textwidth}{@{\extracolsep\fill}lrrrrrrr}
\toprule
~ & ~ & ~ & \multicolumn{2}{c}{95\% CI} & ~ & \multicolumn{2}{c}{OR 95\% CI} \\
\cmidrule{4-5}\cmidrule{7-8}
Parameter & Estimate & Est.\ Error & lower & upper & OR & lower & upper \\
\midrule
Intercept & $-0.08$ & $0.07$ & $-0.21$ & $0.06$ & $0.93$ & $0.81$ & $1.06$ \\
Defaults & $2.89$ & $0.13$ & $2.64$ & $3.15$ & $17.96$ & $13.95$ & $23.44$ \\
Social Influence & $2.48$ & $0.12$ & $2.26$ & $2.72$ & $12.00$ & $9.55$ & $15.21$ \\
Reasoning high & $-0.03$ & $0.10$ & $-0.22$ & $0.15$ & $0.97$ & $0.80$ & $1.17$ \\
Defaults $\times$ Reas.\ high & $-0.23$ & $0.17$ & $-0.57$ & $0.10$ & $0.79$ & $0.57$ & $1.10$ \\
Soc.\ Infl.\ $\times$ Reas.\ high & $0.77$ & $0.18$ & $0.41$ & $1.12$ & $2.15$ & $1.51$ & $3.06$ \\
\midrule
Std.\ dev.\ agent intercept & $0.81$ & $0.05$ & $0.71$ & $0.91$ & -- & -- & -- \\
\botrule
\end{tabular*}
\end{table}

\begin{table}[ht]
\caption{Posterior odds-ratio contrasts for the two per-size pooled models, comprising condition contrasts (H1) and high versus no reasoning contrasts within each condition (H2a and H2b). Entries are posterior medians with 95\% highest-posterior-density intervals. These are marginal contrasts (averaged over reasoning for H1), and therefore differ from the coefficient odds ratios in Tables~\ref{tab:app-sizelarge}--\ref{tab:app-sizesmall}, which condition on the no-reasoning reference level.}\label{tab:app-size-contrasts}
\label{Posterior_size}
\begin{tabular*}{\textwidth}{@{\extracolsep\fill}lcc}
\toprule
Contrast & Flagship pool & Small pool \\
\midrule
Defaults / No nudge & $4.64\ [3.82,\, 5.60]$ & $15.94\ [13.24,\, 19.17]$ \\
Social influence / No nudge & $107.57\ [75.23,\, 147.30]$ & $17.59\ [14.42,\, 20.96]$ \\
Social influence / Defaults & $23.15\ [16.44,\, 31.59]$ & $1.10\ [0.89,\, 1.34]$ \\
\midrule
High / no reasoning, No nudge & $0.94\ [0.71,\, 1.17]$ & $0.97\ [0.80,\, 1.16]$ \\
High / no reasoning, Defaults & $0.39\ [0.29,\, 0.50]$ & $0.77\ [0.56,\, 0.99]$ \\
High / no reasoning, Social influence & $0.85\ [0.44,\, 1.40]$ & $2.09\ [1.50,\, 2.74]$ \\
\botrule
\end{tabular*}
\end{table}

\clearpage

\section{Per-model analyses}\label{app:permodel}

\begin{figure}[ht]
\centering
\begin{subfigure}[b]{0.32\textwidth}
    \centering
    \includegraphics[width=\textwidth]{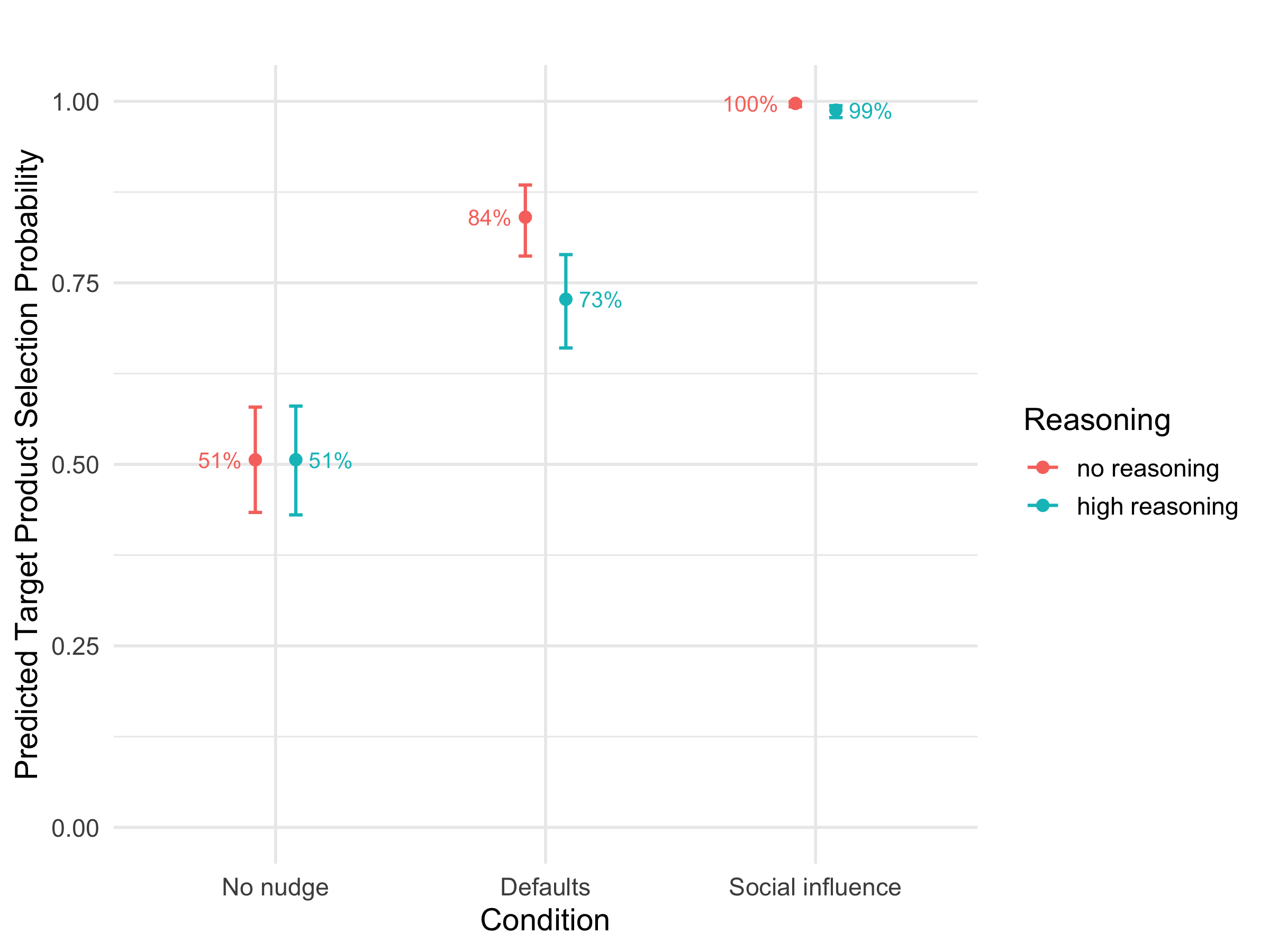}
    \caption{GPT-5.4}
    \label{fig:results_gpt54}
\end{subfigure}
\hfill
\begin{subfigure}[b]{0.32\textwidth}
    \centering
    \includegraphics[width=\textwidth]{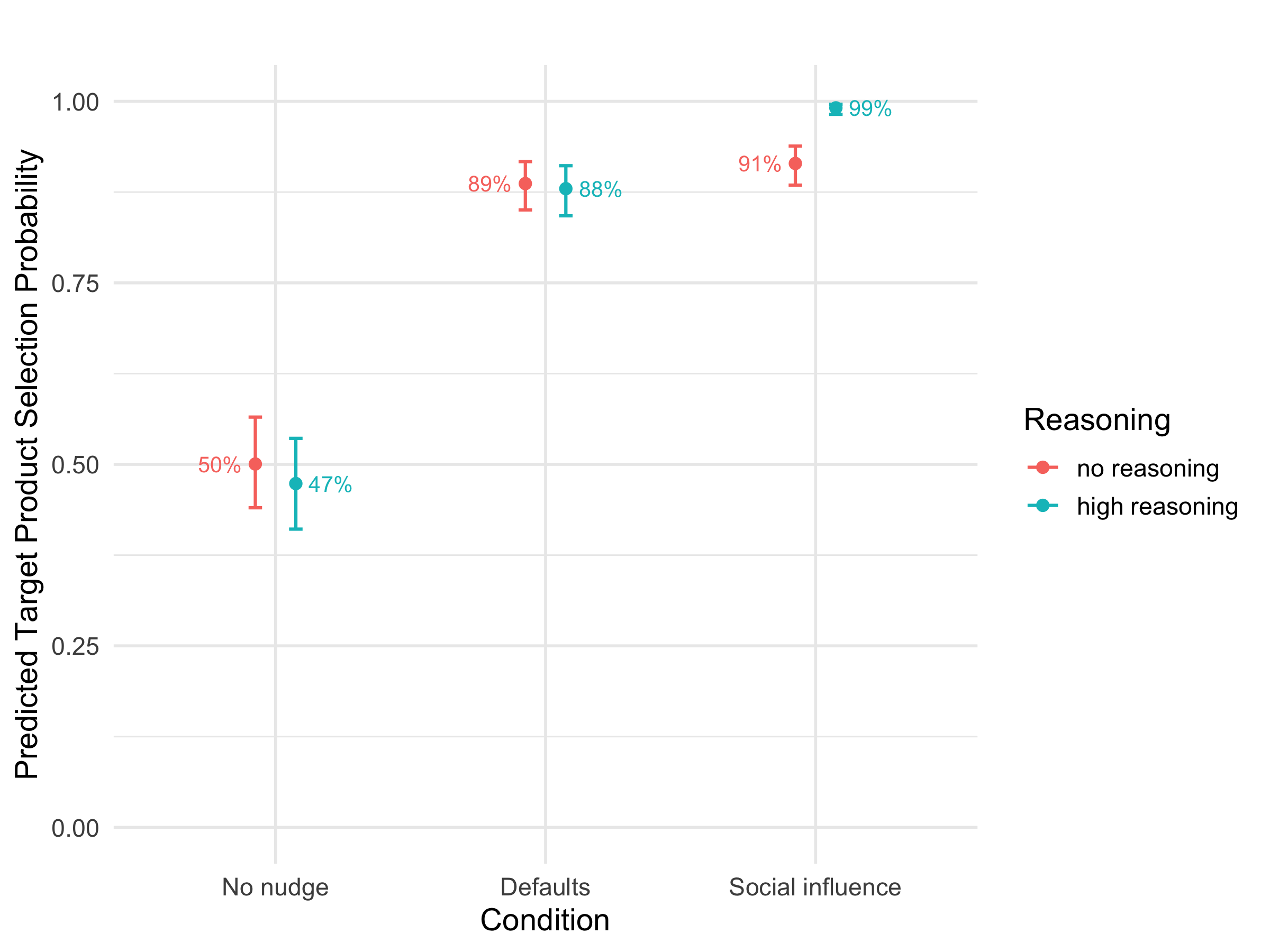}
    \caption{GPT-5.4-mini}
    \label{fig:results_gpt54mini}
\end{subfigure}
\hfill
\begin{subfigure}[b]{0.32\textwidth}
    \centering
    \includegraphics[width=\textwidth]{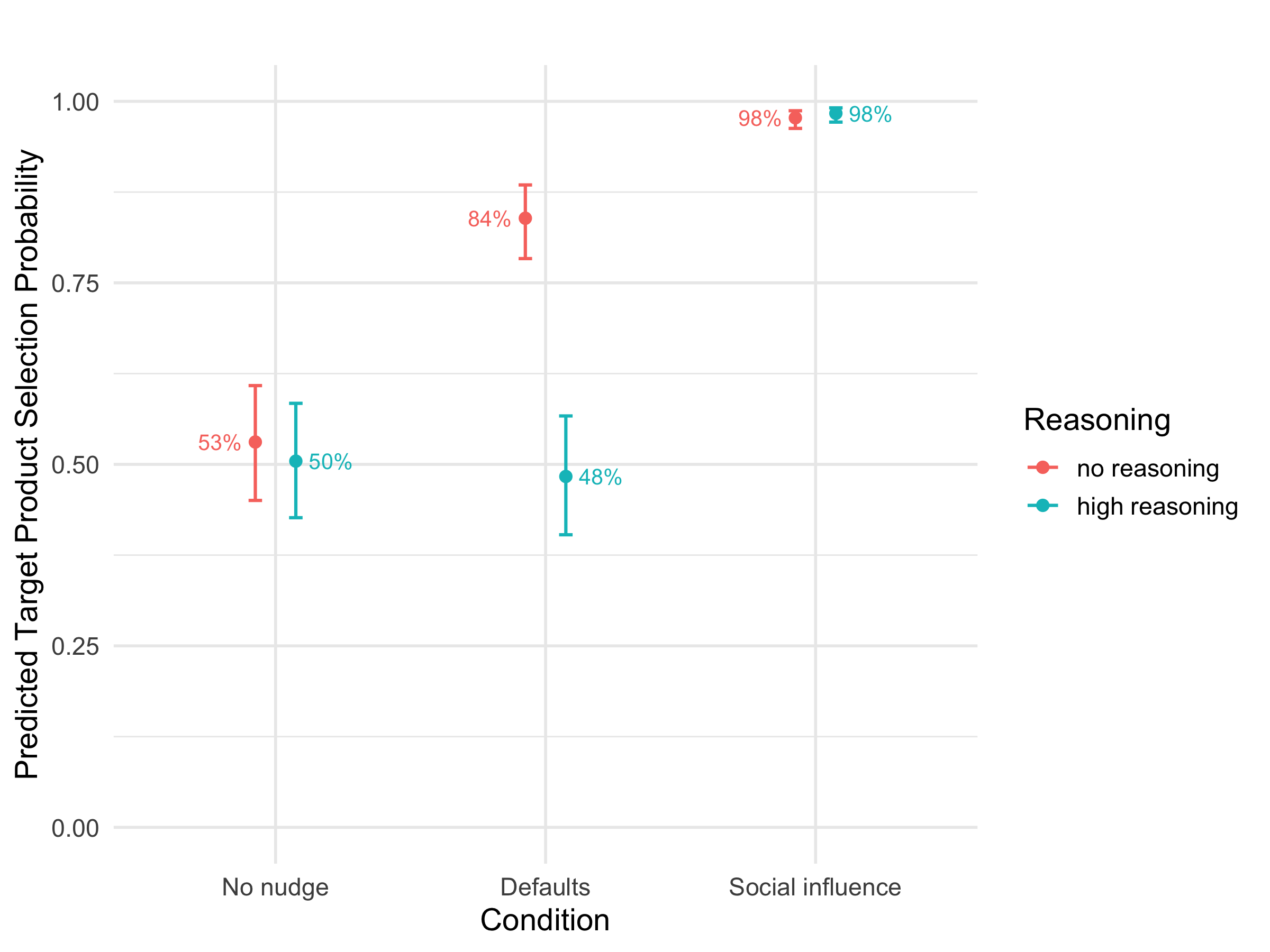}
    \caption{Gemini 3.5 Flash}
    \label{fig:results_g35}
\end{subfigure}
\\[1em]
\begin{subfigure}[b]{0.32\textwidth}
    \centering
    \includegraphics[width=\textwidth]{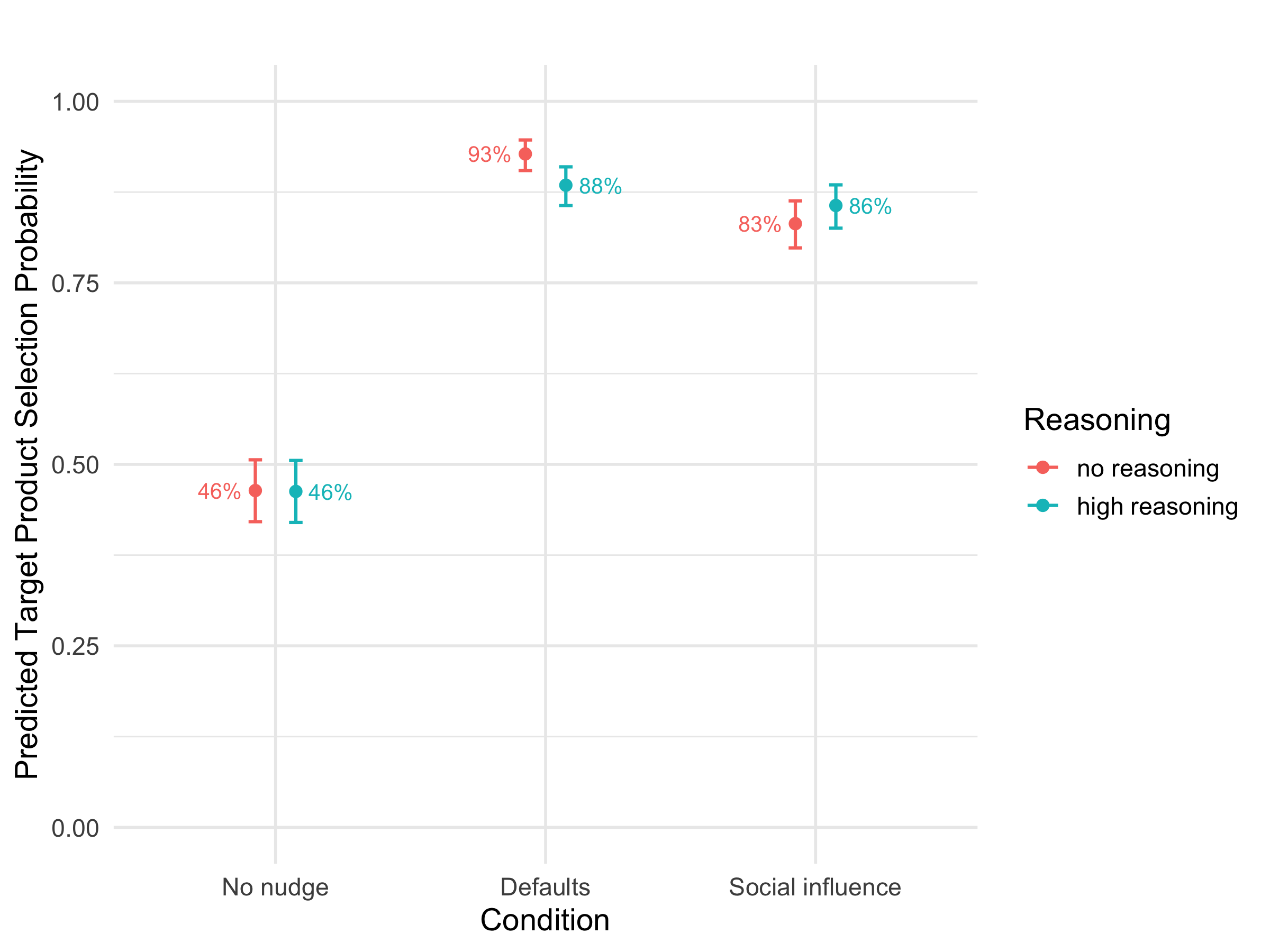}
    \caption{Gemini 3.1 Flash Lite}
    \label{fig:results_g31}
\end{subfigure}
\hfill
\begin{subfigure}[b]{0.32\textwidth}
    \centering
    \includegraphics[width=\textwidth]{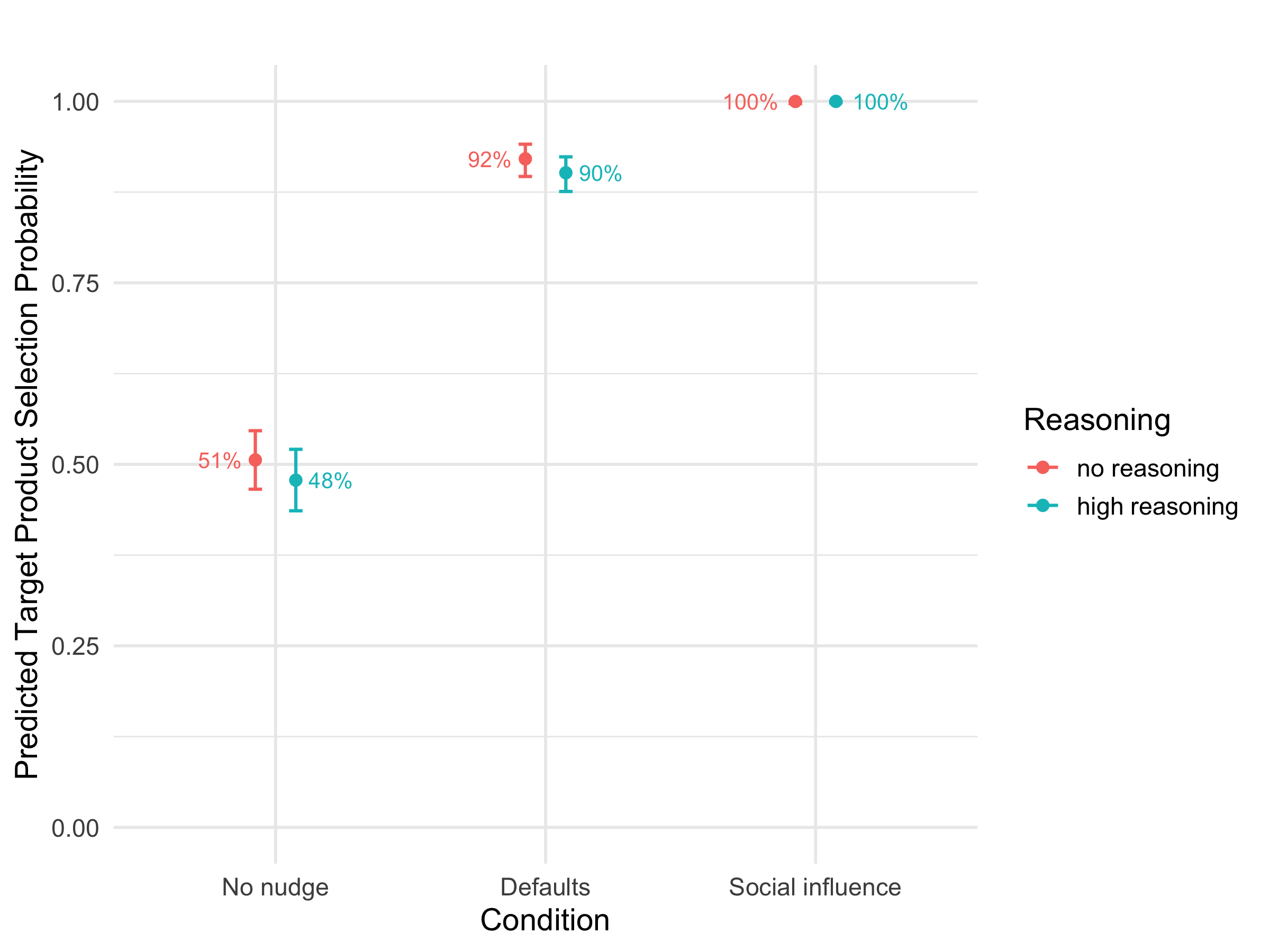}
    \caption{Claude Sonnet 4.6}
    \label{fig:results_sonnet}
\end{subfigure}
\hfill
\begin{subfigure}[b]{0.32\textwidth}
    \centering
    \includegraphics[width=\textwidth]{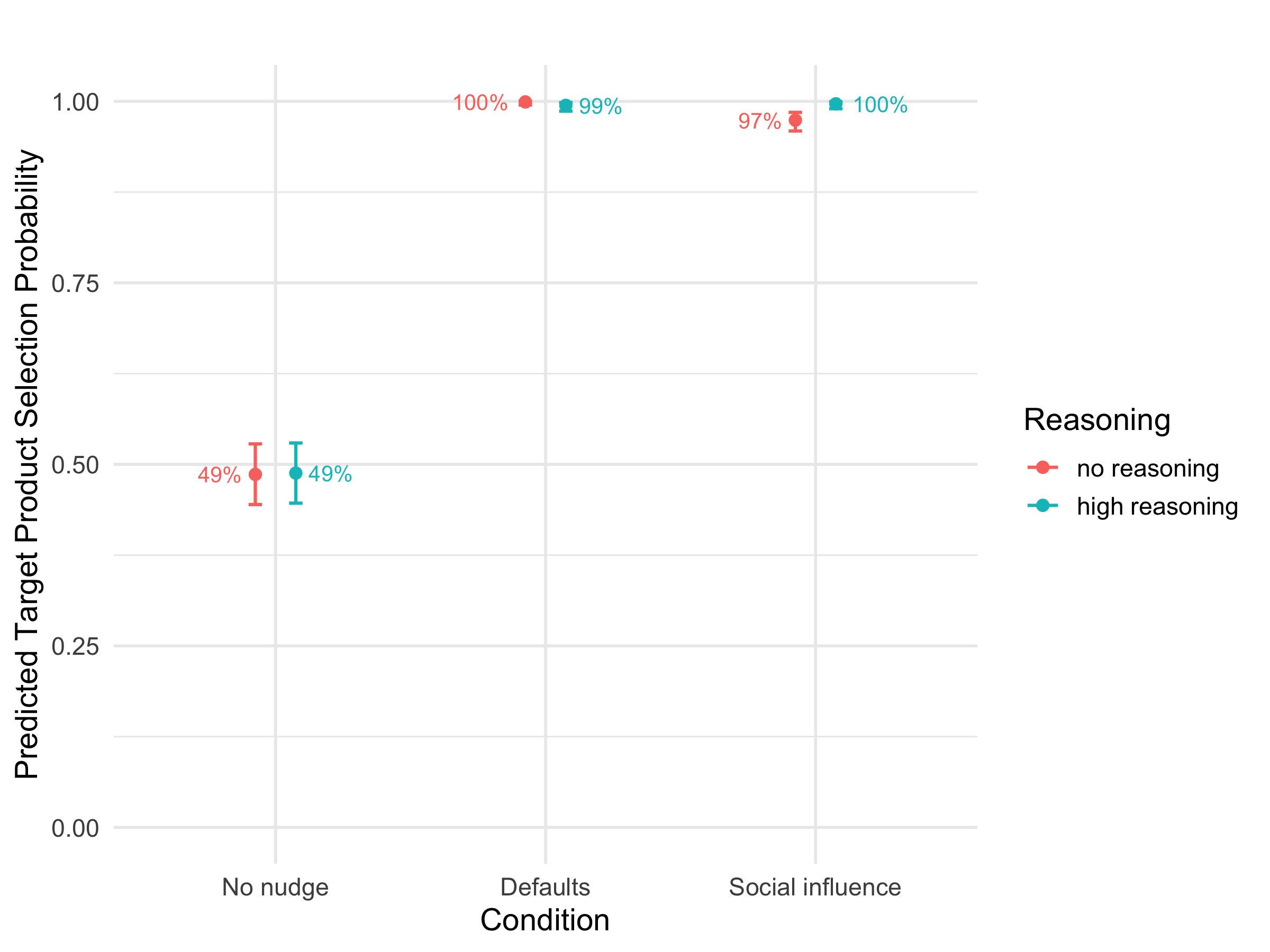}
    \caption{Claude Haiku 4.5}
    \label{fig:results_haiku}
\end{subfigure}
\caption{Predicted target product selection probabilities across conditions and reasoning configurations for the six per-model fits. Red indicates no-reasoning agents, and blue indicates high-reasoning agents.}
\label{fig:results_permodel}
\end{figure}

As a robustness check across individual models, we re-estimated the Model (1) specification separately for each of the six LLM agent backbones ($N = 3{,}600$ observations, $600$ agents each). Tables~\ref{tab:app-gpt54}--\ref{tab:app-haiku} report the complete posterior summaries and Figure~\ref{fig:results_permodel} the corresponding estimated probabilities. Both nudge main effects (H1) are credible in every single model. Note that Claude Sonnet 4.6 exhibits complete separation in the social influence condition (100\% target selection at both reasoning configurations) and Claude Haiku 4.5 quasi-complete separation in the default condition (100\% and 99.3\%). The affected coefficients are partly prior-dependent and should be read via their credible interval bounds.

Regarding the default attenuation under high reasoning (H2a), the effect is concentrated in the flagship models. It is credible for GPT-5.4 ($\beta = -0.68$, $CI_{.95}$ $[-1.30,\, -0.05]$, predicted probability declining from 84\% to 73\%) and strongest for Gemini 3.5 Flash ($\beta = -1.61$, $CI_{.95}$ $[-2.30,\, -0.96]$, declining from 84\% to 48\%, i.e., back to chance level), as well as, more moderately, for Gemini 3.1 Flash Lite ($\beta = -0.51$, $CI_{.95}$ $[-0.96,\, -0.04]$, 93\% to 88\%). By contrast, the smaller GPT-5.4-mini shows virtually no reasoning moderation for defaults ($\beta = 0.04$, $CI_{.95}$ $[-0.51,\, 0.62]$ including $0.0$, 89\% vs.\ 88\%), and neither Claude model yields a credible interaction. Claude Sonnet 4.6 remains highly compliant at both reasoning configurations ($\beta = -0.13$, $CI_{.95}$ $[-0.58,\, 0.33]$, 92\% vs.\ 90\%), and Claude Haiku 4.5 sits at a ceiling in both cells (100\% and 99.3\% raw selection rates, i.e., quasi-complete separation), so its wide interval ($\beta = -2.15$, $CI_{.95}$ $[-5.92,\, 0.21]$) is prior-dependent and uninformative.

Regarding the social influence amplification under high reasoning (H2b), the effect is concentrated in the smaller model variants. It is credible for GPT-5.4-mini ($\beta = 2.46$, $CI_{.95}$ $[1.61,\, 3.38]$, estimated probabilities rising from 91\% to 99\%) and for Claude Haiku 4.5 ($\beta = 2.10$, $CI_{.95}$ $[0.82,\, 3.70]$, rising from 97\% to close to 100\%). For the two Gemini models, the interaction is positive but not credible (Gemini 3.5 Flash $\beta = 0.43$, $CI_{.95}$ $[-0.44,\, 1.30]$, and Gemini 3.1 Flash Lite $\beta = 0.20$, $CI_{.95}$ $[-0.21,\, 0.61]$). For Claude Sonnet 4.6, the social influence condition produced 100\% target selection at both reasoning configurations, leaving no variability for reasoning to modulate, so the resulting interaction estimate is accordingly uninformative ($\beta = 1.11$, $CI_{.95}$ $[-3.42,\, 7.10]$ including $0.0$). Finally, GPT-5.4 constitutes the sole reversal, as its interaction is credibly negative ($\beta = -1.49$, $CI_{.95}$ $[-2.86,\, -0.24]$), although both of its social influence cells lie near the ceiling (99.7\% vs.\ 98.8\% predicted probability), so the practical magnitude of this reversal is small.

\begin{table}[ht]
\caption{GPT-5.4, complete posterior summary.}\label{tab:app-gpt54}
\label{tab:GPT-5.4}
\begin{tabular*}{\textwidth}{@{\extracolsep\fill}lrrrrrrr}
\toprule
~ & ~ & ~ & \multicolumn{2}{c}{95\% CI} & ~ & \multicolumn{2}{c}{OR 95\% CI} \\
\cmidrule{4-5}\cmidrule{7-8}
Parameter & Estimate & Est.\ Error & lower & upper & OR & lower & upper \\
\midrule
Intercept & $0.03$ & $0.15$ & $-0.27$ & $0.32$ & $1.03$ & $0.77$ & $1.37$ \\
Defaults & $1.64$ & $0.24$ & $1.18$ & $2.10$ & $5.14$ & $3.24$ & $8.17$ \\
Social Influence & $5.87$ & $0.60$ & $4.83$ & $7.15$ & $354.15$ & $124.64$ & $1273.98$ \\
Reasoning high & $-0.00$ & $0.22$ & $-0.42$ & $0.42$ & $1.00$ & $0.66$ & $1.52$ \\
Defaults $\times$ Reas.\ high & $-0.68$ & $0.32$ & $-1.30$ & $-0.05$ & $0.51$ & $0.27$ & $0.95$ \\
Soc.\ Infl.\ $\times$ Reas.\ high & $-1.49$ & $0.67$ & $-2.86$ & $-0.24$ & $0.23$ & $0.06$ & $0.79$ \\
\midrule
Std.\ dev.\ agent intercept & $1.23$ & $0.09$ & $1.05$ & $1.42$ & -- & -- & -- \\
\botrule
\end{tabular*}
\end{table}

\begin{table}[ht]
\caption{GPT-5.4-mini, complete posterior summary.}\label{tab:app-gpt54mini}
\label{tab:GPT-5.4-mini}
\begin{tabular*}{\textwidth}{@{\extracolsep\fill}lrrrrrrr}
\toprule
~ & ~ & ~ & \multicolumn{2}{c}{95\% CI} & ~ & \multicolumn{2}{c}{OR 95\% CI} \\
\cmidrule{4-5}\cmidrule{7-8}
Parameter & Estimate & Est.\ Error & lower & upper & OR & lower & upper \\
\midrule
Intercept & $0.00$ & $0.13$ & $-0.24$ & $0.26$ & $1.00$ & $0.79$ & $1.30$ \\
Defaults & $2.06$ & $0.21$ & $1.65$ & $2.48$ & $7.81$ & $5.22$ & $11.90$ \\
Social Influence & $2.37$ & $0.22$ & $1.95$ & $2.80$ & $10.67$ & $7.03$ & $16.38$ \\
Reasoning high & $-0.11$ & $0.18$ & $-0.46$ & $0.24$ & $0.90$ & $0.63$ & $1.27$ \\
Defaults $\times$ Reas.\ high & $0.04$ & $0.29$ & $-0.51$ & $0.62$ & $1.04$ & $0.60$ & $1.85$ \\
Soc.\ Infl.\ $\times$ Reas.\ high & $2.46$ & $0.45$ & $1.61$ & $3.38$ & $11.68$ & $5.02$ & $29.42$ \\
\midrule
Std.\ dev.\ agent intercept & $0.95$ & $0.08$ & $0.78$ & $1.12$ & -- & -- & -- \\
\botrule
\end{tabular*}
\end{table}

\begin{table}[ht]
\caption{Gemini 3.5 Flash, complete posterior summary.}\label{tab:app-g35}
\label{tab:gemini-3-5-flash}
\begin{tabular*}{\textwidth}{@{\extracolsep\fill}lrrrrrrr}
\toprule
~ & ~ & ~ & \multicolumn{2}{c}{95\% CI} & ~ & \multicolumn{2}{c}{OR 95\% CI} \\
\cmidrule{4-5}\cmidrule{7-8}
Parameter & Estimate & Est.\ Error & lower & upper & OR & lower & upper \\
\midrule
Intercept & $0.12$ & $0.16$ & $-0.20$ & $0.44$ & $1.13$ & $0.82$ & $1.55$ \\
Defaults & $1.53$ & $0.25$ & $1.05$ & $2.03$ & $4.62$ & $2.84$ & $7.59$ \\
Social Influence & $3.65$ & $0.32$ & $3.05$ & $4.31$ & $38.43$ & $21.07$ & $74.14$ \\
Reasoning high & $-0.11$ & $0.23$ & $-0.56$ & $0.35$ & $0.90$ & $0.57$ & $1.42$ \\
Defaults $\times$ Reas.\ high & $-1.61$ & $0.34$ & $-2.30$ & $-0.96$ & $0.20$ & $0.10$ & $0.38$ \\
Soc.\ Infl.\ $\times$ Reas.\ high & $0.43$ & $0.44$ & $-0.44$ & $1.30$ & $1.53$ & $0.64$ & $3.66$ \\
\midrule
Std.\ dev.\ agent intercept & $1.36$ & $0.09$ & $1.18$ & $1.54$ & -- & -- & -- \\
\botrule
\end{tabular*}
\end{table}

\begin{table}[ht]
\caption{Gemini 3.1 Flash Lite, complete posterior summary.}\label{tab:app-g31}
\label{tab:gemini-3-1-flash-lite}
\begin{tabular*}{\textwidth}{@{\extracolsep\fill}lrrrrrrr}
\toprule
~ & ~ & ~ & \multicolumn{2}{c}{95\% CI} & ~ & \multicolumn{2}{c}{OR 95\% CI} \\
\cmidrule{4-5}\cmidrule{7-8}
Parameter & Estimate & Est.\ Error & lower & upper & OR & lower & upper \\
\midrule
Intercept & $-0.14$ & $0.09$ & $-0.32$ & $0.03$ & $0.87$ & $0.73$ & $1.03$ \\
Defaults & $2.70$ & $0.18$ & $2.35$ & $3.07$ & $14.84$ & $10.44$ & $21.52$ \\
Social Influence & $1.74$ & $0.15$ & $1.46$ & $2.04$ & $5.71$ & $4.32$ & $7.71$ \\
Reasoning high & $-0.01$ & $0.12$ & $-0.25$ & $0.23$ & $0.99$ & $0.78$ & $1.26$ \\
Defaults $\times$ Reas.\ high & $-0.51$ & $0.24$ & $-0.96$ & $-0.04$ & $0.60$ & $0.38$ & $0.96$ \\
Soc.\ Infl.\ $\times$ Reas.\ high & $0.20$ & $0.21$ & $-0.21$ & $0.61$ & $1.22$ & $0.81$ & $1.84$ \\
\midrule
Std.\ dev.\ agent intercept & $0.30$ & $0.13$ & $0.03$ & $0.54$ & -- & -- & -- \\
\botrule
\end{tabular*}
\end{table}

\begin{table}[ht]

\caption{Claude Sonnet 4.6, complete posterior summary. Complete separation in the social influence condition (100\% target selection at both reasoning configurations), so the affected estimates are partly prior-dependent.}\label{tab:app-sonnet}
\label{tab:Claude Sonnet 4.6}
\begin{tabular*}{\textwidth}{@{\extracolsep\fill}lrrrrrrr}
\toprule
~ & ~ & ~ & \multicolumn{2}{c}{95\% CI} & ~ & \multicolumn{2}{c}{OR 95\% CI} \\
\cmidrule{4-5}\cmidrule{7-8}
Parameter & Estimate & Est.\ Error & lower & upper & OR & lower & upper \\
\midrule
Intercept & $0.02$ & $0.08$ & $-0.14$ & $0.19$ & $1.03$ & $0.87$ & $1.20$ \\
Defaults & $2.43$ & $0.17$ & $2.09$ & $2.78$ & $11.38$ & $8.11$ & $16.17$ \\
Social Influence & $9.31$ & $2.97$ & $5.74$ & $17.15$ & {\footnotesize $11036.99$} & {\footnotesize $310.38$} & {\footnotesize $2.79*10^{7}$}  \\
Reasoning high & $-0.11$ & $0.12$ & $-0.35$ & $0.12$ & $0.89$ & $0.71$ & $1.13$ \\
Defaults $\times$ Reas.\ high & $-0.13$ & $0.24$ & $-0.58$ & $0.33$ & $0.88$ & $0.56$ & $1.40$ \\
Soc.\ Infl.\ $\times$ Reas.\ high & $1.11$ & $2.68$ & $-3.42$ & $7.10$ & $3.05$ & $0.03$ & $1207.72$ \\
\midrule
Std.\ dev.\ agent intercept & $0.16$ & $0.11$ & $0.01$ & $0.40$ & -- & -- & -- \\
\botrule
\end{tabular*}
\end{table}

\begin{table}[ht]
\caption{Claude Haiku 4.5, complete posterior summary. Quasi-complete separation in the default condition (100\% and 99.3\% target selection); the affected estimates are partly prior-dependent.}\label{tab:app-haiku}
\label{tab:Claude Haiku 4.5}
\begin{tabular*}{\textwidth}{@{\extracolsep\fill}lrrrrrrr}
\toprule
~ & ~ & ~ & \multicolumn{2}{c}{95\% CI} & ~ & \multicolumn{2}{c}{OR 95\% CI} \\
\cmidrule{4-5}\cmidrule{7-8}
Parameter & Estimate & Est.\ Error & lower & upper & OR & lower & upper \\
\midrule
Intercept & $-0.06$ & $0.08$ & $-0.22$ & $0.11$ & $0.95$ & $0.80$ & $1.12$ \\
Defaults & $7.42$ & $1.48$ & $5.33$ & $11.15$ & {\footnotesize $1671.81$} & {\footnotesize $206.53$} & {\footnotesize $69250.64$} \\
Social Influence & $3.69$ & $0.27$ & $3.18$ & $4.24$ & $40.07$ & $24.01$ & $69.72$ \\
Reasoning high & $0.01$ & $0.12$ & $-0.23$ & $0.25$ & $1.01$ & $0.79$ & $1.28$ \\
Defaults $\times$ Reas.\ high & $-2.15$ & $1.54$ & $-5.92$ & $0.21$ & $0.12$ & $<0.01$ & $1.23$ \\
Soc.\ Infl.\ $\times$ Reas.\ high & $2.10$ & $0.74$ & $0.82$ & $3.70$ & $8.15$ & $2.27$ & $40.29$ \\
\midrule
Std.\ dev.\ agent intercept & $0.18$ & $0.12$ & $0.01$ & $0.43$ & -- & -- & -- \\
\botrule
\end{tabular*}
\end{table}

\clearpage

\section{Pre-experiment: letter-counting manipulation check}\label{app:reasoningproof}

This appendix reports the design, the statistical tests, and a reasoning-trace analysis for the pre-experiment mentioned in Section~\ref{sec:design}. The pre-experiment grounds the System 1 versus System 2 interpretation of the reasoning configuration. Kahneman treats counting the occurrences of a letter as a canonical System 2 task, one that is effortful, demands sustained attention, and cannot be performed by fast automatic processing \citep{kahneman2011thinking}. Letter counting in LLMs is a direct analogue: it requires inspecting the word character by character rather than retrieving a token-level pattern, and both the architectural analysis of \citet{zhang2024counting} and the empirical work of \citet{xu2025llm} identify the engagement of explicit reasoning, rather than scale or fine-tuning, as the intervention that reliably makes LLMs count letters correctly. If the high-reasoning configuration engages System 2-style deliberation, it should therefore improve letter-counting accuracy. Whether it does, where the improvement is located, and in which output channel the models perform the letter-level work are the subject of this appendix.

\subsection*{Design}

Each of the six model backbones was asked five counting items, 100 times per reasoning configuration, yielding $5 \times 100 \times 2 \times 6 = 6{,}000$ independent API calls ($3{,}000$ per configuration). Items were posed as a bare user message with no system prompt, no browser, and no tools; the provider-side reasoning parameters were the same objects used in the main experiment, imported from the experiment code so that the configurations cannot drift between studies. The five items (Table~\ref{tab:app-rp-items}) comprise one in-distribution word whose answer is plausibly memorized (``raspberry'', the well-known ``strawberry'' case), three novel pseudo-words that cannot be answered from memory, and one zero-count control in which the target letter never appears.

Responses were graded automatically by a case-insensitive word-boundary match on the accepted answers, applied to the \emph{final answer only} and never to the reasoning trace. This is deliberate: traces routinely enumerate the letters correctly (``1\ldots 2\ldots 3'') while the model still emits a wrong final count, so scoring the trace would credit the model for work that did not reach its answer. Word boundaries prevent the digit ``3'' from matching inside ``31'' and ``one'' from matching inside ``none''.

\begin{table}[ht]
\caption{The five counting items. Each was asked 100 times per model and reasoning configuration.}\label{tab:app-rp-items}
\begin{tabular*}{\textwidth}{@{\extracolsep\fill}llcll}
\toprule
Item & Prompt & Count & Accepted answers & Function \\
\midrule
1 & How many \texttt{r}'s in ``raspberry''?   & 3 & three, 3 & In-distribution \\
2 & How many \texttt{w}'s in ``waspbewwy''?   & 3 & three, 3 & Novel pseudo-word \\
3 & How many \texttt{w}'s in ``raspberry''?   & 0 & zero, 0, null, none, no & Zero-count control \\
4 & How many \texttt{s}'s in ``laspbelly''?   & 1 & one, 1 & Novel pseudo-word \\
5 & How many \texttt{b}'s in ``baspbebbby''?  & 5 & five, 5 & Novel pseudo-word \\
\botrule
\end{tabular*}
\end{table}

\subsection*{Accuracy and statistical tests}

\begin{figure}[ht]
\centering
\includegraphics[width=.5\textwidth]{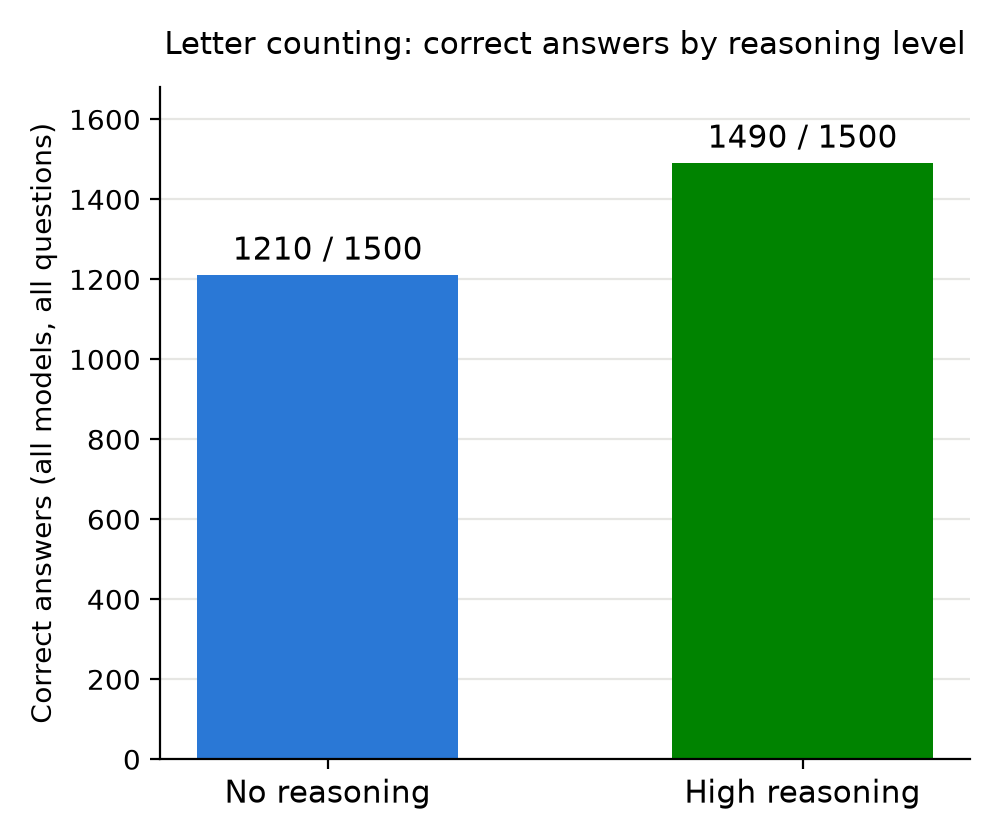}
\caption{Correct answers on the letter-counting task, pooled across all six models and five items (3{,}000 calls per configuration), under the no-reasoning and high-reasoning configurations.}\label{fig:manipulation}
\end{figure}

Pooled across all models and items, accuracy rose from $80.3\%$ ($2{,}410/3{,}000$) under no reasoning to $99.3\%$ ($2{,}980/3{,}000$) under high reasoning, a risk difference of $19.0$ percentage points (Figure~\ref{fig:manipulation}). The gain is concentrated exactly where the theory predicts, namely on the novel pseudo-words that cannot be answered from memory, and not at all on the control item (Table~\ref{tab:app-rp-items-acc}). No model was ever less accurate under high reasoning, on any item (Table~\ref{tab:app-rp-cells}).

\begin{table}[ht]
\caption{Correct answers per item, pooled over the six models (600 calls per item and configuration).}\label{tab:app-rp-items-acc}
\begin{tabular*}{\textwidth}{@{\extracolsep\fill}llrr}
\toprule
Item & Type & No reasoning & High reasoning \\
\midrule
1 \ ``raspberry'' / \texttt{r} & Memorizable & $491/600\ (81.8\%)$ & $600/600\ (100.0\%)$ \\
2 \ ``waspbewwy'' / \texttt{w} & Novel & $318/600\ (53.0\%)$ & $581/600\ (96.8\%)$ \\
3 \ ``raspberry'' / \texttt{w} & Zero-count control & $600/600\ (100.0\%)$ & $600/600\ (100.0\%)$ \\
4 \ ``laspbelly'' / \texttt{s} & Novel & $467/600\ (77.8\%)$ & $600/600\ (100.0\%)$ \\
5 \ ``baspbebbby'' / \texttt{b} & Novel & $534/600\ (89.0\%)$ & $599/600\ (99.8\%)$ \\
\midrule
Total & & $2{,}410/3{,}000\ (80.3\%)$ & $2{,}980/3{,}000\ (99.3\%)$ \\
\botrule
\end{tabular*}
\end{table}

The five items differ enormously in baseline difficulty, from a floor of $53\%$ on item~2 to a ceiling on item~3, so a single comparison of all $500$ calls per configuration would confound the reasoning effect with item composition. We therefore tested the reasoning effect with the Cochran--Mantel--Haenszel (CMH) test \citep{cochran1954some,Rayner2020,mantel1959statistical}, which is \emph{stratified} by item. The two configurations are compared within each item separately, and the five within-item comparisons are then pooled into a single test statistic. The procedure is in effect a fixed-effect meta-analysis across items, in which every item serves as its own control. The model-level tests are Holm-corrected over the four testable models, since the two Claude models admit no test (see below). Significance is assessed by randomization rather than by the $\chi^2$ approximation. The reasoning labels are reshuffled within each item ($20{,}000$ resamples), and the $p$-value is the proportion of reshuffles that produce an association at least as strong as the observed one. This procedure is exact for sparse and well-filled items alike, which is why we applied it uniformly instead of switching tests where cells are small. A data-dependent choice of test would amount to selecting the test after seeing the data. 

Two features of the data govern how Table~\ref{tab:app-rp-tests} should be read. The first is \emph{ceiling}. In 18 of the 30 model~$\times$~item cells, both configurations are at $100\%$. Such cells carry no information about the reasoning effect and drop out of the CMH statistic automatically. Both Claude models are at $500/500$ in \emph{both} configurations, so they exhibit no variance at all and are reported as untestable rather than as null results. The second is \emph{separation}. In cells such as GPT-5.4 on item~2 which is correct in 0 of 100 calls without reasoning and in 100 of 100 with it (Table~\ref{tab:app-rp-cells}, one configuration is perfect and the other at zero, which makes the odds ratio infinite. The CMH test itself remains well-defined in this situation, but the odds ratio does not, so the effect size to read is the risk difference, that is, the improvement in percentage points. Because every stratum contains exactly 100 calls per configuration, the marginal risk difference equals the unweighted mean of the per-item risk differences.

\begin{table}[ht]
\caption{Letter-counting accuracy per model and the reasoning effect. $\Delta$ (pp) is the improvement in percentage points under high reasoning. $p$-values are CMH tests stratified by item, assessed by randomization ($20{,}000$ within-item reshuffles, bounded below by $1/20{,}001$) and Holm-corrected across the four testable models. The two Claude models are at $100\%$ in both configurations and therefore admit no test.}\label{tab:app-rp-tests}
\begin{tabular*}{\textwidth}{@{\extracolsep\fill}llrrrl}
\toprule
Model & Size & No reasoning & High reasoning & $\Delta$ (pp) & $p$ (Holm) \\
\midrule
GPT-5.4               & Flagship & $384/500\ (76.8\%)$ & $500/500\ (100.0\%)$ & $+23.2$ & $<0.001$ \\
GPT-5.4-mini          & Small    & $163/500\ (32.6\%)$ & $480/500\ (96.0\%)$  & $+63.4$ & $<0.001$ \\
Gemini 3.5 Flash      & Flagship & $493/500\ (98.6\%)$ & $500/500\ (100.0\%)$ & $+1.4$  & $0.014$ \\
Gemini 3.1 Flash Lite & Small    & $370/500\ (74.0\%)$ & $500/500\ (100.0\%)$ & $+26.0$ & $<0.001$ \\
Claude Sonnet 4.6     & Flagship & $500/500\ (100.0\%)$ & $500/500\ (100.0\%)$ & $0.0$ & untestable \\
Claude Haiku 4.5      & Small    & $500/500\ (100.0\%)$ & $500/500\ (100.0\%)$ & $0.0$ & untestable \\
\botrule
\end{tabular*}
\end{table}

\begin{table}[ht]
\caption{Correct answers per model and item (out of 100 calls per cell) under the no-reasoning (No) and high-reasoning (High) configurations. Items as defined in Table~\ref{tab:app-rp-items}. This is the cell-level grid underlying the two marginal tables above.}\label{tab:app-rp-cells}
\begin{tabular*}{\textwidth}{@{\extracolsep\fill}lrrrrrrrrrr}
\toprule
~ & \multicolumn{2}{c}{Item 1} & \multicolumn{2}{c}{Item 2} & \multicolumn{2}{c}{Item 3} & \multicolumn{2}{c}{Item 4} & \multicolumn{2}{c}{Item 5} \\
\cmidrule{2-3}\cmidrule{4-5}\cmidrule{6-7}\cmidrule{8-9}\cmidrule{10-11}
Model & No & High & No & High & No & High & No & High & No & High \\
\midrule
GPT-5.4               & $95$  & $100$ & $0$   & $100$ & $100$ & $100$ & $99$  & $100$ & $90$  & $100$ \\
GPT-5.4-mini          & $5$   & $100$ & $9$   & $81$  & $100$ & $100$ & $5$   & $100$ & $44$  & $99$ \\
Gemini 3.5 Flash      & $93$  & $100$ & $100$ & $100$ & $100$ & $100$ & $100$ & $100$ & $100$ & $100$ \\
Gemini 3.1 Flash Lite & $98$  & $100$ & $9$   & $100$ & $100$ & $100$ & $63$  & $100$ & $100$ & $100$ \\
Claude Sonnet 4.6     & $100$ & $100$ & $100$ & $100$ & $100$ & $100$ & $100$ & $100$ & $100$ & $100$ \\
Claude Haiku 4.5      & $100$ & $100$ & $100$ & $100$ & $100$ & $100$ & $100$ & $100$ & $100$ & $100$ \\
\botrule
\end{tabular*}
\end{table}

\begin{figure}[ht]
\centering
\includegraphics[width=.7\textwidth]{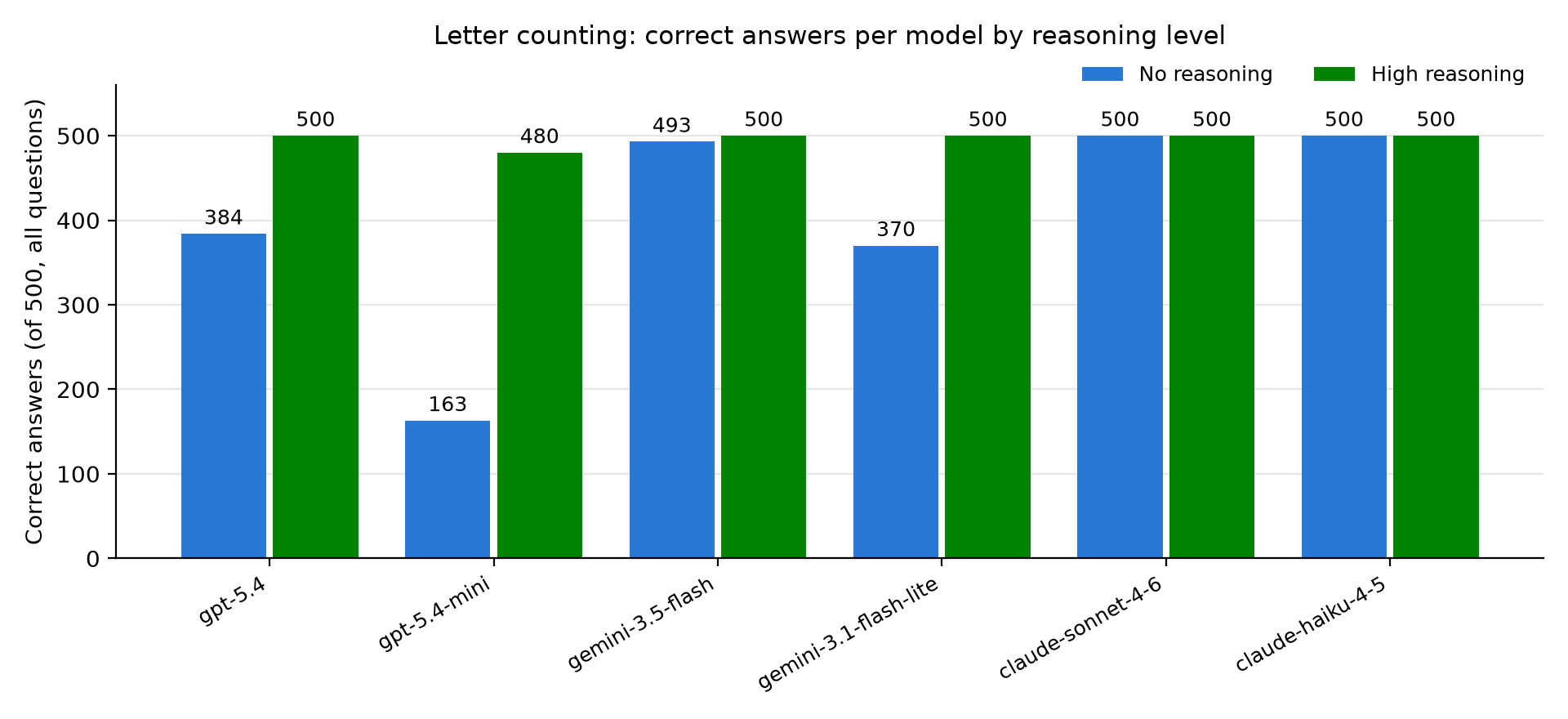}
\caption{Correct answers on the letter-counting task per model backbone, pooled over the five items (500 calls per model and configuration), under the no-reasoning and high-reasoning configurations.}\label{fig:app-rp-permodel}
\end{figure}

At the aggregate level the effect is credible under every test we considered. The CMH test stratified by all 30 model~$\times$~item cells yields a pooled (Mantel--Haenszel) odds ratio of $266.1$, the same quantity a fixed-effect meta-analysis would report, with a randomization $p < 0.001$. Because the individual calls are clustered within cells, we additionally report two tests that compress each model~$\times$~item cell into a single paired observation and are therefore immune to both the ceiling and the clustering. Of the 30 cells, 12 improved under high reasoning, none deteriorated, and 18 were tied at the ceiling, giving an exact sign test $p = 0.0005$ and a Wilcoxon signed-rank $p = 0.0022$. We note for completeness that a single unstratified $2 \times 2$ test over all $6{,}000$ calls would return $p < 10^{-100}$, but such a test treats 100 correlated calls per cell as independent observations and overstates the evidence by orders of magnitude. It is not the basis for any claim here. Figure~\ref{fig:app-rp-permodel} shows the per-model accuracies.

\subsection*{Reasoning-trace analysis}

Accuracy establishes \emph{that} the manipulation works, but not \emph{how}. We therefore annotated all $6{,}000$ responses for whether the text decomposes the word into individual letters, that is, whether it spells the word out (``r-a-s-p-b-e-r-r-y''), marks the target letters inside the word, enumerates the letters as list items, or walks their positions or indices. Merely \emph{talking about} decomposing (``I will go through each letter'') was scored negative, as the construct of interest is the observable decomposition itself. The annotation was applied separately to two channels: the model's visible answer, and its hidden reasoning trace where one was returned.

Because a single automatic labeller would be unvalidated, we used two independent ones. The first is a deterministic pattern matcher covering the notational variants observed in the corpus. The second is a panel of three LLM judges, one per provider (Gemini 3.1 Flash Lite, GPT-5.4-mini, Claude Haiku 4.5), each returning a structured verdict under a fixed rubric and blind to which model produced the text, with the majority vote taken as the panel label; a panel of small, diverse judges evaluates more cheaply and with less intra-model bias than a single large judge \citep{verga2024replacing}. The two labellers agree almost perfectly on the visible answer (Cohen's $\kappa = 0.996$, raw agreement $99.8\%$, $n = 6{,}000$), and the three judges agree among themselves at Fleiss' $\kappa = 0.987$, with $5{,}944$ of $6{,}000$ verdicts unanimous. Agreement on the trace channel is high in raw terms ($96.0\%$) but the corresponding $\kappa$ is depressed to $0.721$ by the strong prevalence skew ($93\%$ of traces are positive); the free-marginal (Randolph) coefficient, which is robust to that skew, is $0.885$. Because the three judges are themselves three of the six annotated backbones, we verified that this cannot have driven the labels: recomputing the panel for each provider's own $2{,}000$ rows while excluding that provider's judge reproduces the full panel's label on every row on which the reduced two-judge panel is decisive (agreement $=1.000$; the remaining 2, 8, and 15 rows per provider are ties between the two remaining judges and admit no verdict).

\begin{table}[ht]
\caption{Percentage of calls in which the model decomposes the word into individual letters, by channel (500 calls per model and configuration). ``Visible'' refers to the model's answer text, ``trace'' to its hidden reasoning trace, and ``any channel'' to either. Traces exist only under the high-reasoning configuration; 8 GPT-5.4 and 3 GPT-5.4-mini high-reasoning calls returned no trace, so their trace percentages are based on 492 and 497 calls, while the any-channel column counts those calls as non-decomposing (which is why it can fall below the trace column).}\label{tab:app-rp-decomp}
\begin{tabular*}{\textwidth}{@{\extracolsep\fill}lrrrrr}
\toprule
~ & \multicolumn{2}{c}{Visible answer} & Trace & \multicolumn{2}{c}{Any channel} \\
\cmidrule{2-3}\cmidrule{5-6}
Model & No reas. & High reas. & High reas. & No reas. & High reas. \\
\midrule
GPT-5.4               & $5.2$   & $0.0$   & $90.0$  & $5.2$   & $88.6$ \\
GPT-5.4-mini          & $0.4$   & $0.0$   & $76.9$  & $0.4$   & $76.4$ \\
Gemini 3.5 Flash      & $96.0$  & $91.6$  & $92.4$  & $96.0$  & $99.6$ \\
Gemini 3.1 Flash Lite & $67.8$  & $59.2$  & $98.4$  & $67.8$  & $99.6$ \\
Claude Sonnet 4.6     & $87.0$  & $100.0$ & $100.0$ & $87.0$  & $100.0$ \\
Claude Haiku 4.5      & $100.0$ & $87.4$  & $100.0$ & $100.0$ & $100.0$ \\
\midrule
Pooled                & $59.4$  & $56.4$  & $93.0$  & $59.4$  & $94.0$ \\
\botrule
\end{tabular*}
\end{table}

Table~\ref{tab:app-rp-decomp} shows that the reasoning configuration does \emph{not} make models decompose more visibly: pooled, the rate is $59.4\%$ without reasoning and $56.4\%$ with it. What changes is the channel. Under high reasoning every model decomposes the word in its hidden trace on the large majority of calls ($76.9$--$100\%$), so that decomposition somewhere in the observable output rises from $59.4\%$ to $94.0\%$. The provider-reported token counts corroborate this directly for the four models that expose them: the median number of reasoning tokens under the no-reasoning configuration is $0$ in every case, rising under high reasoning to $88.5$ (GPT-5.4), $65$ (GPT-5.4-mini), $278$ (Gemini 3.5 Flash) and $283.5$ (Gemini 3.1 Flash Lite). Anthropic does not report a separate reasoning-token count, so no token-level figure can be given for the two Claude models; their trace column rests on the returned thinking text alone.

The decisive pattern is the relationship between the two analyses. A model's accuracy gain from the reasoning toggle is almost perfectly inversely related to how often it already decomposes the word \emph{without} reasoning enabled:

\begin{itemize}
\item The two \textbf{Claude} models decompose the word in essentially every no-reasoning answer ($100\%$ and $87\%$), spelling it out and enumerating positions in the visible reply even though the thinking channel is disabled. They are consequently already at $100\%$ accuracy without reasoning, and the toggle has no room left to work ($\Delta = 0$ for both).
\item The two \textbf{OpenAI} models essentially never decompose visibly ($5.2\%$ and $0.4\%$, falling to $0\%$ when reasoning is enabled and the work moves into the trace). They answer in a single terse sentence and depend entirely on the hidden channel, and they show the largest accuracy gains ($+23.2$ and $+63.4$ percentage points).
\item The two \textbf{Google} models are intermediate, decomposing visibly in $68$--$96\%$ of no-reasoning answers, with correspondingly intermediate gains ($+26.0$ and $+1.4$ points).
\end{itemize}

Two implications follow for the interpretation of the main experiment. First, the manipulation is best understood as suppressing the reasoning \emph{channel} rather than as switching deliberation off outright: a model whose training disposes it to externalize step-by-step work in its answer will continue to do so under the no-reasoning configuration. Second, and consequently, the manipulation is not equally strong across providers. It is strongest for the OpenAI backbones, intermediate for Google, and weakest for Anthropic, where the no-reasoning configuration still yields visibly deliberative behavior. This asymmetry is a plausible contributor to the per-model heterogeneity in the reasoning moderation reported in the per-model appendix of the main article, and it suggests that a null reasoning effect for the Claude models should be read as a restriction of range in the manipulation rather than as evidence that deliberation does not matter for nudge susceptibility. It also indicates that ``reasoning off'' is not a provider-independent construct, which is a caveat for any study that treats a vendor's reasoning switch as a uniform experimental factor.

We note two limitations. First, the decomposition measure is a surface-form property of the observable text and is not itself a measure of internal computation; a model could in principle perform character-level work without externalizing it in either channel. Second, the trace channel is only as observable as the provider makes it: the returned thinking content may be a summary rather than the full chain of thought, and Anthropic exposes no reasoning-token count, so the channel evidence is corroborated by token accounting for the OpenAI and Google models but rests on the returned thinking text alone for the two Claude models.

\clearpage

\end{appendices}

\end{document}